\documentclass[10pt,twocolumn,letterpaper]{article}

\usepackage{cvpr}
\usepackage{times}
\usepackage{epsfig}
\usepackage{graphicx}
\usepackage{amsmath}
\usepackage{amssymb}
\usepackage{algorithm}
\usepackage{algpseudocode}
\usepackage{bm}
\usepackage{mathtools}
\DeclarePairedDelimiter{\floor}{\lfloor}{\rfloor}
\usepackage{array, multirow}
\usepackage{changepage}

\usepackage[breaklinks=true,bookmarks=false,colorlinks]{hyperref}

\cvprfinalcopy 

\def\cvprPaperID{6770} 

\begin{document}

\title{Unbiased Scene Graph Generation from Biased Training}

\author{
    Kaihua Tang\textsuperscript{1},\quad Yulei Niu\textsuperscript{3},\quad Jianqiang Huang\textsuperscript{1,2}\thanks{Corresponding author.},\quad Jiaxin Shi\textsuperscript{4},\quad Hanwang Zhang\textsuperscript{1}\\
    {\small \textsuperscript{1}Nanyang Technological University,\quad \textsuperscript{2}Damo Academy, Alibaba Group,\quad \textsuperscript{3}Renmin University of China,\quad \textsuperscript{4}Tsinghua University}\\
    {\tt\small kaihua001@e.ntu.edu.sg,\quad niu@ruc.edu.cn,\quad jianqiang.jqh@gmail.com}\\
    {\tt\small shijx12@163.com,\quad hanwangzhang@ntu.edu.sg}
}

\maketitle


\begin{abstract}
Today's scene graph generation (SGG) task is still far from practical, mainly due to the severe training bias, e.g., collapsing diverse \texttt{human walk on/ sit on/lay on beach} into \texttt{human on beach}. Given such SGG, the down-stream tasks such as VQA can hardly infer better scene structures than merely a bag of objects. However, debiasing in SGG is not trivial because traditional debiasing methods cannot distinguish between the good and bad bias, e.g., good context prior (e.g., \texttt{person read book} rather than \texttt{eat}) and bad long-tailed bias (e.g., \texttt{near} dominating \texttt{behind/in front of}). In this paper, we present a novel SGG framework based on \textbf{causal inference} but not the conventional likelihood. We first build a causal graph for SGG, and perform traditional biased training with the graph. Then, we propose to draw the \textbf{counterfactual causality} from the trained graph to infer the effect from the bad bias, which should be removed. In particular, we use \textbf{Total Direct Effect} as the proposed final predicate score for unbiased SGG. Note that our framework is agnostic to any SGG model and thus can be widely applied in the community who seeks unbiased predictions. By using the proposed \textbf{Scene Graph Diagnosis} toolkit\footnote{Our code is publicly available on GitHub: \url{https://github.com/KaihuaTang/Scene-Graph-Benchmark.pytorch}} on the SGG benchmark Visual Genome and several prevailing models, we observed significant improvements over the previous state-of-the-art methods.
\end{abstract}

\section{Introduction}
\label{sec:intro}

\begin{figure}[t]
   \begin{minipage}[b]{1.0\linewidth}
   \centerline{\includegraphics[width=80mm]{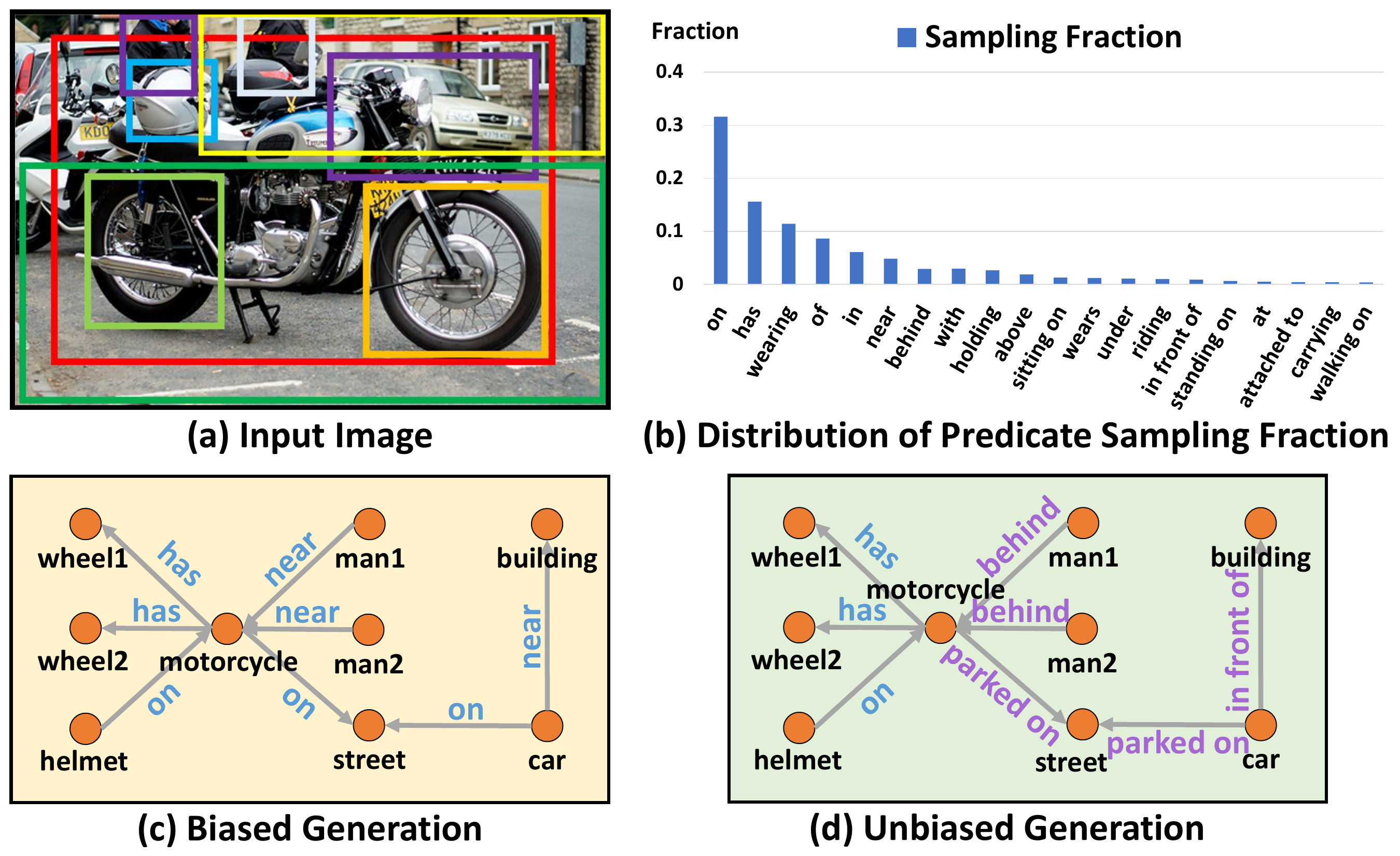}}
   \end{minipage}
   \caption{An example of scene graph generation (SGG). (a) An input image with bounding boxes. (b) The distribution of sample fraction for the most frequent 20 predicates in Visual Genome~\cite{krishna2017visual}. (c) SGG from re-implemented MOTIFS~\cite{zellers2018neural}. (d) SGG by the proposed unbiased prediction from the same model.}
   \label{fig:1} 
   \vspace{-0.2in}
\end{figure}

Scene graph generation (SGG)~\cite{xu2017scene} --- a visual detection task of objects and their relationships in an image --- seems to have never fulfilled its promise: a comprehensive visual scene representation that supports \emph{graph reasoning} for high-level tasks such as visual captioning~\cite{yao2018exploring, yang2019auto} and VQA~\cite{teney2017graph, hudson2019gqa}. Once equipped with SGG, these high-level tasks have to abandon the ambiguous visual relationships --- yet on which are our core efforts made~\cite{zellers2018neural,tang2019learning,chen2019knowledge}, then pretend that there is a graph --- nothing but a sparse object layout with binary links, and finally shroud it into graph neural networks~\cite{yan2018spatial} for merely more contextual object representations~\cite{yang2019auto, johnson2018image, teney2017graph}. Although this is partly due to the research gap in graph reasoning~\cite{battaglia2018relational, shi2019explainable, hudson2019learning}, the crux lies in the \emph{biased} relationship prediction.

Figure~\ref{fig:1} visualizes the SGG results from a state-of-the-art model~\cite{zellers2018neural}. We can see a frustrating scene: among almost perfectly detected objects, most of their visual relationships are trivial and less informative. For example in Figure~\ref{fig:1}(c), except the trivial 2D spatial layouts, we know little about the image from \texttt{near}, \texttt{on}, and \texttt{has}. Such heavily biased generation comes from the \emph{biased training data}, more specifically, as shown in Figure~\ref{fig:1}(b), the highly-skewed long-tailed relationship annotations. For example, if a model is trained for predicting \texttt{on} 1,000 times more than \texttt{standing on}, then, during test, the former is more likely to prevail over the latter. Therefore, to perform a sensible graph reasoning, we need to distinguish more fine-grained relationships from the ostensibly probable but trivial ones, such as replacing \texttt{near} with \texttt{behind/in front of}, and \texttt{on} with \texttt{parking on/driving on} in Figure~\ref{fig:1}(d). 


However, we should not blame the biased training because both our visual world \textit{per se} and the way we describe it are biased: there are indeed more \texttt{person carry bag} than \texttt{dog carry bag} (\ie, the long-tail theory); it is easier for us to label \texttt{person beside table} rather than \texttt{eating on} (\ie, bounded rationality~\cite{simon1990bounded}); and we prefer to say \texttt{person on bike} rather than \texttt{person ride on bike} (\ie, language or reporting bias~\cite{misra2016seeing}). In fact, most of the biased annotations can help the model learn good contextual prior~\cite{lu2016visual,zellers2018neural} to filter out the unnecessary search candidates such as \texttt{apple park on table} and \texttt{apple wear hat}. A promising but embarrassing finding~\cite{zellers2018neural} is that: by only using the statistical prior of detected object class in the Visual Genome benchmark~\cite{krishna2017visual}, we can already achieved 30.1\% on Recall@100 for Scene Graph Detection --- rendering all the much more complex SGG models almost useless --- that is only 1.1-1.5\% lower than the state-of-the-art~\cite{chen2019counterfactual, tang2019learning, zhang2019graphical}. Not surprisingly, as we will show in Section~\ref{sec:exp}, conventional debiasing methods who do not respect the ``good bias'' during training, \eg, re-sampling~\cite{he2009learning} and re-weighting~\cite{lin2017focal}, fail to generalize to unseen relationships, \ie, zero-shot SGG~\cite{lu2016visual}.

\begin{figure}[t!]
   \begin{minipage}[b]{1\linewidth}
   \centerline{\includegraphics[width=80mm]{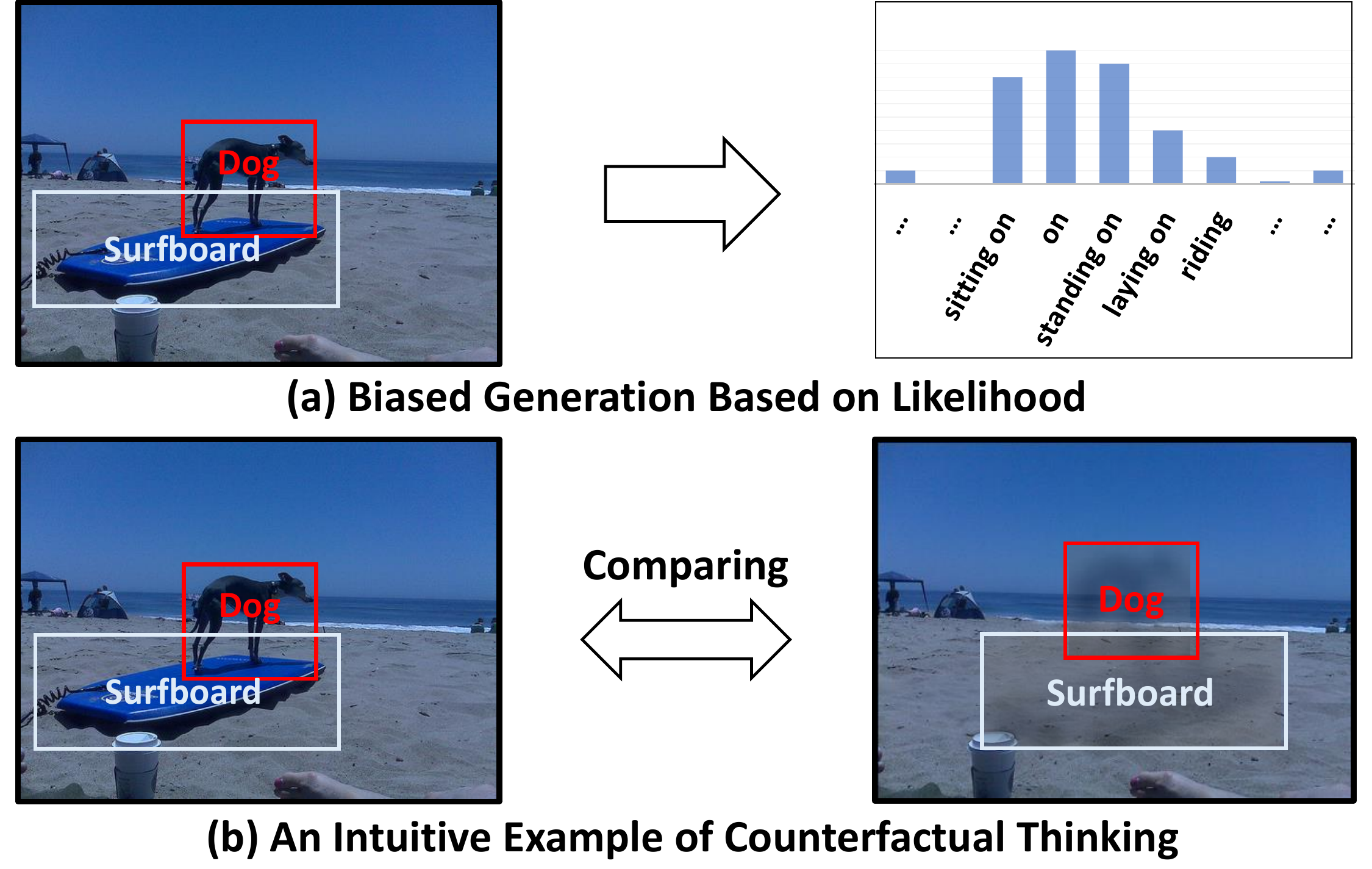}}
   \end{minipage}
   \caption{(a) The biased generation that directly predicts labels from likelihood. (b) An intuitive example of the proposed total direct effect, which calculates the difference between the real scene and the counterfactual one. Note that the ``wipe-out'' is only for the illustrative purpose but not considered as visual processing.}
   \label{fig:2} 
   \vspace{-0.2in}
\end{figure}

For both machines and humans, decision making is a collaboration of \emph{content} (endogenous reasons) and \emph{context} (exogenous reasons)~\cite{van2015cognitive}. Take SGG as an example, in most SGG models~\cite{zellers2018neural,chen2019counterfactual,zhang2019graphical}, the content is the visual features of the subject and object, and the context is the visual features of the subject-object union regions and the pairwise object classes. We humans --- born and raised in the biased nature --- are ambidextrous in embracing the good while avoiding the bad context, and making unbiased decisions together with the content. The underlying mechanism is \emph{causality-based}: the decision is made by pursuing the main causal effect caused by the content but not the side-effect by context. However, on the other hand, machines are usually \emph{likelihood-based}: the prediction is analogous to look-up the content and its context in a huge likelihood table, interpolated by population training. We believe that the key is to teach machines how to distinguish between the ``main effect'' and ``side-effect''. 

In this paper, we propose to empower machines the ability of \emph{counterfactual causality}~\cite{Judea2018thebookofwhy} to pursue the ``main effect'' in unbiased prediction:
\begin{adjustwidth}{0.3cm}{0.3cm}
\vspace{1pt}
\noindent\textit{If I \textbf{had not} seen the content, would I still make the same prediction?}
\vspace{1pt}
\end{adjustwidth}
The counterfactual lies between the fact that ``I see'' and the imagination ``I had not'', and the comparison between the factual and counterfactual will naturally \emph{remove} the effect from the context bias, because the context is the only thing unchanged between the two alternatives.

To better illustrate the profound yet subtle difference between likelihood and counterfactual causality, we present a \texttt{dog standing on surfboard} example in Figure~\ref{fig:2}(a). Due to the biased training, the model will eventually predict the \texttt{on}. Note that even though the rest choices are not all exactly correct, thanks to the bias, they still help to filter out a large amount of unreasonable ones. To take a closer look at what relationship it is in the context bias, we are essentially comparing the original scene with a counterfactual scene (Figure~\ref{fig:2}(b)): only the visual features of the \texttt{dog} and \texttt{surfboard} are wiped out, while keeping the rest --- the scene and the object classes --- untouched, as if the visual features had ever existed. By doing this, we can focus on the main visual effects of the relationship without losing the context. 

\begin{figure}[t!]
   \begin{minipage}[b]{1.0\linewidth}
   \centerline{\includegraphics[width=80mm]{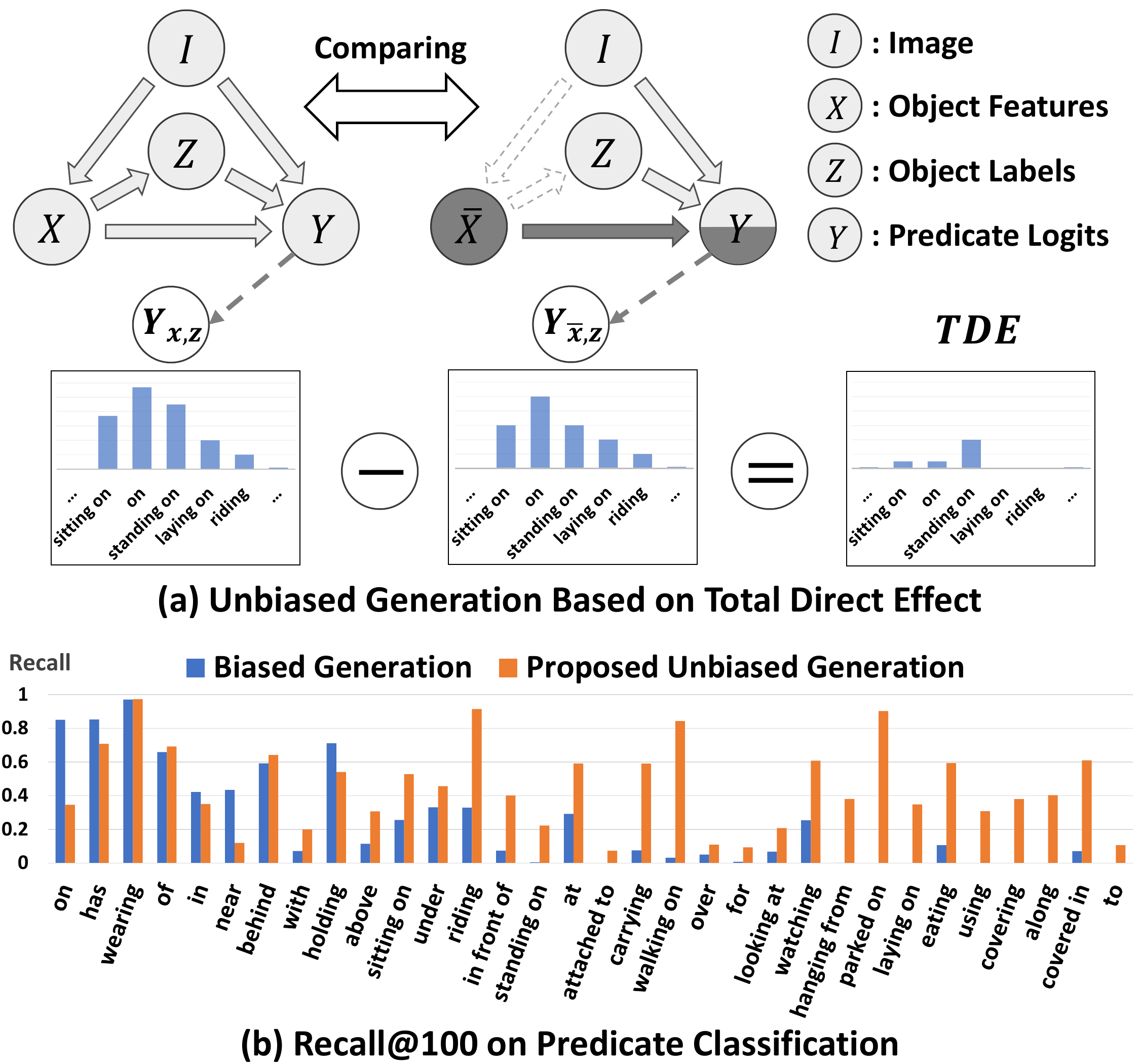}}
   \end{minipage}
   \caption{(a) The example of total direct effect calculation and corresponding operations on the causal graph, where $\bar{X}$ represents wiped-out $X$. (b) Recall@100 of Predicate Classification for selected predicates ranking by sampling fraction. The biased generation refers to re-implemented MOTIFS~\cite{zellers2018neural} and the proposed unbiased generation is the result from the same model using TDE.}
   \label{fig:3} 
   \vspace{-0.2in}
\end{figure}

We propose a novel unbiased SGG method based on the Total Direct Effect (TDE) analysis framework in causal inference~\cite{vanderweele2015explanation, pearl2001direct, vanderweele2013three}. Figure~\ref{fig:3}(a) shows the underlying causal graphs~\cite{pearl2016causal, Judea2018thebookofwhy} of the two alternate scenes: factual and counterfactual. Although a formal introduction of them is given in Section~\ref{sec:biased_training}-\ref{sec:unbiased_pred}, now you can simply understand the nodes as data features and the directed links as (parametric) data flows. For example, $X\rightarrow Y$, $Z\rightarrow Y$, and $I\rightarrow Y$ indicate that the relationship $Y$ is a combined effect caused by \emph{content}: the pair of object visual features $X$, \emph{context}: their object classes $Z$, and \emph{scene}: the image $I$; the faded links denote that the wiped-out $\bar{X}$ is no longer caused by $I$ or affects $Z$. These graphs offer an algorithmic formulation to calculate TDE, which exactly realizes the counterfactual thinking in Figure~\ref{fig:2}. As shown in Figure~\ref{fig:3}(b), the proposed TDE significantly improves most of the predicates, and impressively, the distribution of the improved performances is no longer long-tailed, indicating the fact that our improvement is indeed from the proposed method, but NOT from the better exploitation of the context bias. A closer analysis in Figure~\ref{fig:5} further shows that the worse predictions like \texttt{on} --- though very few --- are due to turning to more fine-grained results such as \texttt{stand on} and \texttt{park on}. We highlight that TDE is a model-agnostic prediction strategy and thus applicable for a variety of models and fusion tricks~\cite{zhang2017visual, zellers2018neural, tang2019learning}.

Last but not least, we propose a new standard of SGG diagnosis toolkit (cf. Section~\ref{sec:metrics}) for more comprehensive SGG evaluations. Besides traditional evaluation tasks, it consists of the bias-sensitive metric: mean Recall~\cite{tang2019learning, chen2019knowledge} and a new Sentence-to-Graph Retrieval for a more comprehensive graph-level metric. By using this toolkit on SGG benchmark Visual Genome~\cite{krishna2017visual} and several prevailing baselines, we verify the severe bias in existing models and demonstrate the effectiveness of the proposed unbiased prediction over other debiasing strategies.

\section{Related Work}
\noindent\textbf{Scene Graph Generation.} SGG~\cite{xu2017scene, zellers2018neural} has received increasing attention in computer vision community, due to the potential revolution that would be brought to down-stream visual reasoning tasks~\cite{shi2019explainable, yang2019auto, krishna2018referring, johnson2018image}. Most of the existing methods~\cite{xu2017scene, woo2018linknet, dai2017detecting, li2017scene, yin2018zoom, tang2019learning, yang2018graph, gu2019scene, qi2019attentive, wang2019exploring} struggle for better feature extraction networks. Zellers~\etal~\cite{zellers2018neural} firstly brought the bias problem of SGG into attention and the followers~\cite{tang2019learning, chen2019knowledge} proposed the unbiased metric (mean Recall), yet, their approaches are still restricted to the feature extraction networks, leaving the biased SGG problem unsolved. The most related work \cite{liang2019vrr} just prunes those dominant and easy-to-predict relationships in the training set.
 

\noindent\textbf{Unbiased Training.} The bias problem has long been investigated in machine learning~\cite{torralba2011unbiased}. Existing debiasing methods can be roughly categorized into three types: 1) data augmentation or re-sampling~\cite{geirhos2018imagenettrained, li2018resound, li2019repair, he2009learning, burnaev2015influence}, 2) unbiased learning through elaborately designed training curriculums or learning losses~\cite{zemel2013learning, lin2017focal}, 3) disentangling biased representations from the unbiased~\cite{misra2016seeing, cadene2019rubi}. The proposed TDE analysis can be regarded as the third category, but the main difference is that TDE doesn't require to train additional layers like \cite{misra2016seeing, cadene2019rubi} to model the bias, it directly separates the bias from existing models through the counterfactual surgeries on causal graphs.

\noindent\textbf{Mediation Analysis.} It is also known as effect analysis~\cite{vanderweele2015explanation, Judea2018thebookofwhy}, which is widely adopted in medical, political or psychological research~\cite{richiardi2013mediation, keele2015statistics, dunn2015evaluation, mackinnon2007mediation, king2008political} as the tool of studying the effect of certain treatments or policies. However, it has been neglected in the community of computer vision for years. There are very few recent works~\cite{nair2019causal, kusner2017counterfactual, niu2020counterfactual, qi2019two, wang2020visual, he2020learning, yang2020deconfounded} trying to endow the model with the capability of causal reasoning. More detailed background knowledge can be found in \cite{pearl2016causal,Judea2018thebookofwhy, vanderweele2015explanation}.

\begin{figure}[t]
   \begin{minipage}[b]{1.0\linewidth}
   \centerline{\includegraphics[width=80mm]{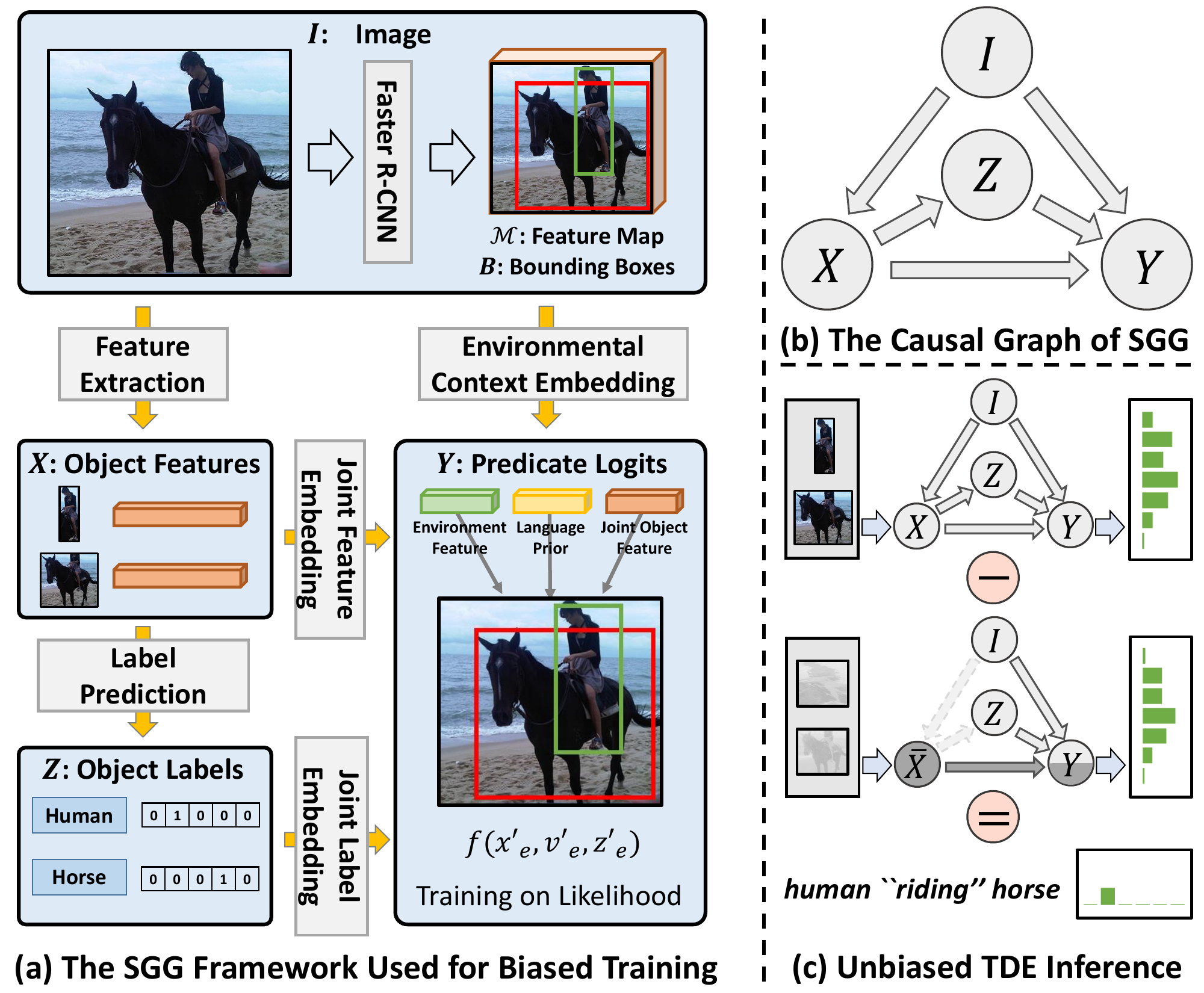}}
   \end{minipage}
   \caption{(a) The framework used in our biased training. (b) The causal graph of the SGG framework. (c) An illustration of the proposed TDE inference.}
   \label{fig:4} 
   \vspace{-0.2in}
\end{figure}

\section{Biased Training Models in Causal Graph}
\label{sec:biased_training}
As illustrated in Figure~\ref{fig:4}, we summarize the SGG framework in the form of \textit{Causal Graph} (\textit{a.k.a.}, structural causal model)~\cite{Judea2018thebookofwhy, pearl2000causality, pearl2016causal}. It is a directed acyclic graph $\mathcal{G}=\{\mathcal{N}, \mathcal{E}\}$, indicating how a set of variables $\mathcal{N}$ interact with each other through the causal links $\mathcal{E}$. It provides a sketch of the causal relations behind the data and how variables obtain their values, \eg, $(I,X,Z)\to Y$. Before we conduct counterfactual analysis that deliberately manipulates the values of nodes and prunes the causal graph, we first revisit the conventional biased SGG model training in the graphical view. 

The causal graph in Figure~\ref{fig:4}(b) is applicable to a variety of SGG methods, since it is highly general, imposing no constraints on the detailed implementations. We case-study three representative model formulations: the classic VTransE~\cite{zhang2017visual}, the state-of-the-art MOTIFS~\cite{zellers2018neural} and VCTree~\cite{tang2019learning}, using the language of nodes and links.

\noindent\textbf{Node} $\bm{I}$ (\textbf{Input Image\&Backbone}). A Faster R-CNN~\cite{ren2015faster} is pre-trained and frozen in this node, It outputs a set of bounding boxes $B=\{b_i|i=1...n\}$ and the feature map $\mathcal{M}$ from image $I$.

\noindent\textbf{Link} $\bm{I\to X}$ (\textbf{Object Feature Extractor}). It firstly extracts RoIAlign features~\cite{he2017mask} $R=\{r_i\}$ and tentative object labels $L=\{l_i\}$ by the object classifier on Faster R-CNN. Then, like MOTIFS~\cite{zellers2018neural} or VCTree~\cite{tang2019learning}, we can use the following module to encode visual contexts for each object:
\begin{equation}
    \textit{Input}: \{(r_i, b_i, l_i)\} \Longrightarrow \textit{Output}: \{x_i\},
    \label{module:1}
\end{equation}
where MOTIFS implements it as bidirectional LSTMs (Bi-LSTMs) and VCTree~\cite{tang2019learning} adopts bidirectional TreeLSTMs (Bi-TreeLSTMs)~\cite{tai2015improved}, early works like VTransE~\cite{zhang2017visual} simply use fully connected layers. 

\noindent\textbf{Node} $\bm{X}$ (\textbf{Object Feature}). The pairwise object feature $X$ takes value from $\{(x_i, x_j)|i\neq j;i,j=1...n\}$. We slightly abuse the notation hereinafter, denoting the combination of representations from $i$ and $j$ as subscript $e$: $x_{e}=(x_i,x_j)$.

\noindent\textbf{Link} $\bm{X\to Z}$ (\textbf{Object Classification}). The fine-tuned label of each object is decoded from the corresponding $x_i$ by: 
\begin{equation}
    \textit{Input}: \{x_i\} \Longrightarrow \textit{Output}: \{z_i\},
    \label{module:2}
\end{equation}
where MOTIFS~\cite{zellers2018neural} and VCTree~\cite{tang2019learning} utilizes LSTM and TreeLSTM as decoders to capture the co-occurrence among object labels, respectively. The input of each LSTM/ TreeLSTM cell is the concatenation of feature and the previous label $[x_i;z_{i-1}]$. VTransE~\cite{zhang2017visual} uses the conventional fully connected layer as the classifier.

\noindent\textbf{Node} $\bm{Z}$ (\textbf{Object Class}). It contains a pair of one-hot vectors for object labels $z_{e}=(z_i, z_j)$.

\noindent\textbf{Link} $\bm{X\to Y}$ (\textbf{Object Feature Input for SGG}). For relationship classification, pairwise feature $X$ are merged into a joint representation by the module:
\begin{equation}
    \textit{Input}: \{x_{e}\} \Longrightarrow \textit{Output}: \{x'_{e}\},
    \label{module:3}
\end{equation}
where another Bi-LSTMs and Bi-TreeLSTMs layers are applied in MOTIFS~\cite{zellers2018neural} and VCTree~\cite{tang2019learning}, respectively, before concatenating the pair of object features. VTransE~\cite{zhang2017visual} uses fully connected layers and element-wise subtraction for feature merging.

\noindent\textbf{Link} $\bm{Z\to Y}$ (\textbf{Object Class Input for SGG}). The language prior is calculated in this link through a joint embedding layer $z'_{e}=W_z[z_i \otimes z_j]$, where $\otimes$ generates the one-hot unique vector $\mathbb{R}^{N\times N}$ for the pair of $N$-way object labels.

\noindent\textbf{Link} $\bm{I\to Y}$ (\textbf{Visual Context Input for SGG}). This link extracts the contextual union region features $v'_{e}=\text{Convs}(\text{RoIAlign}(\mathcal{M},b_i \cup b_j))$ where $b_i \cup b_j$ indicates the union box of two RoIs.

\noindent\textbf{Node} $\bm{Y}$ (\textbf{Predicate Classification}). The final predicate logits $Y$ that takes inputs from the three branches is then generated by using a fusion function. In Section~\ref{sec:exp}, we test two general fusion functions: 1) SUM: $y_{e}=W_x x'_{e} + W_v v'_{e} + z'_{e}$, 2) GATE:  $y_{e}= W_r x'_{e} \cdot \sigma(W_x x'_{e} + W_v v'_{e} + z'_{e})$, where $\cdot$ is element-wise product, $\sigma(\cdot)$ is a sigmoid function.

\noindent\textbf{Training Loss}. All models are trained by using the conventional cross-entropy losses of object labels and predicate labels. To avoid any single link spontaneously dominating the generation of logits $y_{e}$, especially $Z\to Y$, we further add auxiliary cross-entropy losses that individually predict $y_{e}$ from each branch.

\section{Unbiased Prediction by Causal Effects}
\label{sec:unbiased_pred}
Once the above training has been done, the causal dependencies among the variables are learned, in terms of the model parameters. The conventional biased prediction can only see the output of the entire graph given an image $I=u$ without any idea about how a specific pair of objects affect their predicate. However, causal inference~\cite{Judea2018thebookofwhy} encourages us to think out of the black box. From the graphical point of view, we are no longer required to run the entire graph as a whole. We can directly manipulate the values of several nodes and see what would be going on. For example, we can cut off the link $I\to X$ and assign a dummy value to $X$, then investigate what the predicate would be. The above operation is termed \emph{intervention} in causal inference~\cite{pearl2016causal}. Next, we will make unbiased predictions by intervention and its induced counterfactuals.

\vspace{-0.05in}
\subsection{Notations}
\begin{figure}[t]
   \begin{minipage}[b]{1.0\linewidth}
   \centerline{\includegraphics[width=70mm]{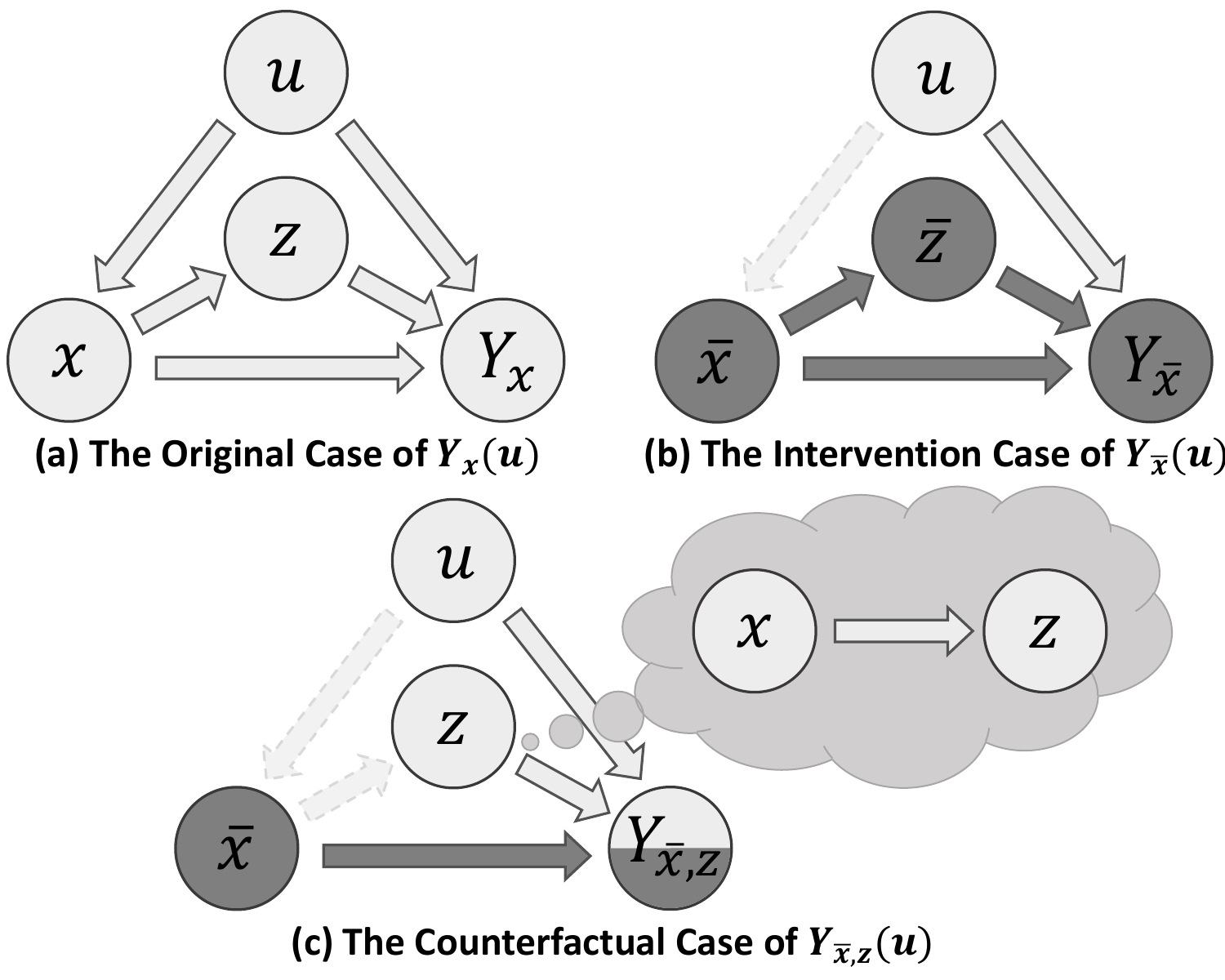}}
   \end{minipage}
   \caption{The original causal graph of SGG together with two interventional and counterfactual alternates.}
   \label{fig:counterfactual} 
   \vspace{-0.2in}
\end{figure}

\vspace{-0.05in}
\noindent\textbf{Intervention}. It can be denoted as $\bm {do(\cdot)}$. It wipes out all the in-coming links of a variable and demands the variable to take a certain value, \eg $do(X=\bar{x})$ in Figure~\ref{fig:counterfactual}(b), meaning $X$ is no longer affected by its causal parents. 

\noindent\textbf{Counterfactual}. It means \textit{``counter to the facts''}~\cite{roese1997counterfactual}, and takes one step further that assigns the ``clash of worlds'' combination of values to variables. Take Figure~\ref{fig:counterfactual}(c) as an example, if the intervention $do(X=\bar{x})$ is conducted on $X$, the variable $Z$ still takes the original $z$ as if $x$ had existed.


\noindent\textbf{Causal Effect}. Throughout this section, we will use the pairwise object feature $X$ as our control variable where the intervention is conducted, aiming to assess its effects, due to the fact that there wouldn't be any valid relationship if the pair of objects do not exist. The observed $X$ is denoted as $x$ while the intervened unseen value is $\bar{x}$, which is set to either the mean feature of the training set or zero vector. The object label $z$ on Figure~\ref{fig:counterfactual}(c) is calculated from Eq.~(\ref{module:2}), taking $x$ as input. We denote the output logits $Y$ after the intervention $X=\bar{x}$ as follows (Figure~\ref{fig:counterfactual}(b)):
\begin{equation}
    Y_{\bar{x}}(u)=Y(do(X=\bar{x})|u),
\end{equation}
where $u$ is the input image in SGG. Following the above notation, the original and counterfactual $Y$, \ie, Figure~\ref{fig:counterfactual}(a,c), can be  re-written as $Y_x(u)$ and $Y_{\bar{x},z}(u)$, respectively.

\vspace{-0.05in}
\subsection{Total Direct Effect}
\vspace{-0.05in}
As we discussed in Section~\ref{sec:intro}, instead of the static likelihood that tends to be biased, the unbiased prediction lies in the difference between the observed outcome $Y_x(u)$ and its counterfactual alternate $Y_{\bar{x},z}(u)$. The later one is a context-specific bias that we want to remove from prediction. Intuitively, the unbiased prediction that we seek is the visual stimuli from blank to the observed real objects with specific attributes, states, and behaviors, but not merely from the surroundings and language priors. Those specific visual cues of objects are the key to the more fine-grained and informative unbiased predictions, because even if the overall prediction is biased towards the relationship like \texttt{dog on surfboard}, the ``straight legs'' would cause more effect on \texttt{standing on} rather than \texttt{sitting on}. In causal inference~\cite{vanderweele2015explanation, vanderweele2013three}, the above prediction process can be calculated as Total Direct Effect (TDE):
\begin{equation}
    TDE = Y_{x}(u) - Y_{\bar{x},z}(u),
    \label{eq:tde}
\end{equation}
where the first term is from the original graph and the second one is from the counterfactual, as illustrated in Figure~\ref{fig:counterfactual}. 

Note that there is another type of effect~\cite{vanderweele2015explanation}, Total Effect (TE), which is easy to be mixed up with TDE. Instead of deriving counterfactual bias $Y_{\bar{x},z}(u)$, TE lets all the descendant nodes of $X$ change with intervention $do(X=\bar{x})$ as shown in Figure~\ref{fig:counterfactual}(b). TE is therefore formulated as:
\begin{equation}
    TE = Y_{x}(u) - Y_{\bar{x}}(u).
    \label{eq:te}
\end{equation}
The main difference lies in the fact that $Y_{\bar{x}}(u)$ is not conditioned on the original object labels (those caused by $x$), so TE only removes the general bias in the whole dataset (similar to the $b$ in $y = k\cdot x +b$), rather than the specific bias caused by the mediator we care about. The subtle difference between TE and TDE is further defined as Natural Indirect Effect (NIE)~\cite{vanderweele2015explanation} or Pure Indirect Effect (PIE)~\cite{vanderweele2013three}. More experimental analyses among these three types of effect are given in Section~\ref{sec:exp}.

\noindent\textbf{Overall SGG.} At last, the proposed unbiased prediction $y^{\dagger}_{e}$ is obtained by replacing the conventional one-time prediction with TDE, which essentially ``thinks'' twice: one for observational $Y_{x_{e}}(u)=y_{e}$, the other for imaginary $Y_{\bar{x},z_e}(u) = y_{e}(\bar{x}, z_{e})$. The unbiased logits of Y is therefore defined as follows:
\begin{equation}
    y^{\dagger}_{e} = y_{e} - y_{e}(\bar{x}, z_{e}).
\end{equation}
It is also worth mentioning that the proposed TDE doesn't introduce any additional parameters and is widely applicable to a variety of models.

\section{Experiments}
\label{sec:exp}
\vspace{-0.05in}
\subsection{Settings and Models}
\vspace{-0.05in}
\noindent\textbf{Dataset.} For SGG, we used Visual Genome (VG)~\cite{krishna2017visual} dataset to train and evaluate our models, which is composed of 108k images across 75k object categories and 37k predicate categories. However, as 92\% of the predicates have no more than 10 instances, we followed the widely adopted VG split~\cite{xu2017scene, zellers2018neural, tang2019learning, chen2019counterfactual} containing the most frequent 150 object categories and 50 predicate categories. The original split only has training set (70\%) and test set (30\%). We followed \cite{zellers2018neural} to sample a 5k validation set from training set for parameter tuning. For Sentence-to-Graph Retrieval (cf. Section~\ref{sec:metrics}), we selected the overlapped 41,859 images between VG and MS-COCO Caption dataset~\cite{lin2014microsoft} and divided them into train/test-1k/test-5k (35,859/1,000/5,000) sets. The later two only contain images from VG test set in case of exposing to grount-truth SGs. Each image has at least 5 captions serving as human queries, the same as how we use searching engines.

\noindent\textbf{Model Zoo.} We evaluated three models: VTransE~\cite{zhang2017visual}, MOTIFS~\cite{zellers2018neural}, VTree~\cite{tang2019learning}, and two fusion functions: SUM and GATE. They were re-implemented using the same codebase as we proposed. All models shared the same hyper-parameters and the pre-trained detector backbone.

\begin{table*}
\centering
\scalebox{0.8}{
\begin{tabular}{r | r | r | c c c |c c c|c c c}
\hline
\multicolumn{3}{c}{} & \multicolumn{3}{c}{Predicate Classification} & \multicolumn{3}{c}{Scene Graph Classification} & \multicolumn{3}{c}{Scene Graph Detection} \\
\hline
Model & Fusion & Method & mR@20 & mR@50 & mR100 & mR@20 & mR50 & mR100 & mR@20 & mR50 & mR100  \\ 
\hline 
IMP+~\cite{xu2017scene, chen2019knowledge} & - & - & - & 9.8 & 10.5 & - & 5.8 & 6.0 & - & 3.8 & 4.8  \\
FREQ~\cite{zellers2018neural, tang2019learning} & - & - & 8.3 & 13.0 & 16.0 & 5.1 & 7.2 & 8.5 & 4.5 & 6.1 & 7.1 \\
MOTIFS~\cite{zellers2018neural, tang2019learning} & - & - & 10.8 & 14.0 & 15.3 & 6.3 & 7.7 & 8.2 & 4.2 & 5.7 & 6.6  \\
KERN~\cite{chen2019knowledge} & - & - & - & 17.7 & 19.2 & - & 9.4 & 10.0 & - & 6.4 & 7.3  \\
VCTree~\cite{tang2019learning} & - & - & 14.0 & 17.9 & 19.4 & 8.2 & 10.1 & 10.8 & 5.2 & 6.9 & 8.0 \\
\hline 
\multirow{12}*{MOTIFS\textsuperscript{$\dagger$}} & \multirow{10}*{SUM} & Baseline & 11.5 & 14.6 & 15.8 & 6.5 & 8.0 & 8.5 & 4.1 & 5.5 & 6.8 \\ 
\cline{3-12}
& & Focal & 10.9 & 13.9 & 15.0 & 6.3 & 7.7 & 8.3 & 3.9 & 5.3 & 6.6 \\
& & Reweight & 16.0 & 20.0 & 21.9 & 8.4 & 10.1 & 10.9 & \textbf{6.5} & \textbf{8.4} & \textbf{9.8} \\
& & Resample & 14.7 & 18.5 & 20.0 & 9.1 & 11.0 & 11.8 & 5.9 & 8.2 & 9.7 \\
\cline{3-12}
& & X2Y & 13.0 & 16.4 & 17.6 & 6.9 & 8.6 & 9.2 & 5.1 & 6.9 & 8.1 \\
& & X2Y-Tr & 11.6 & 14.9 & 16.0 & 6.5 & 8.4 & 9.1 & 5.0 & 6.9 & 8.1 \\
\cline{3-12}
& & TE & 18.2 & 25.3 & 29.0 & 8.1 & 12.0 & 14.0 & 5.7 & 8.0 & 9.6 \\
& & NIE & 0.6 & 1.1 & 1.4 & 6.1 & 9.0 & 10.6 & 3.8 & 5.1 & 6.0\\
& & TDE & \textbf{18.5} & \textbf{25.5} & \textbf{29.1} & \textbf{9.8} & \textbf{13.1} & \textbf{14.9} & 5.8 & 8.2 & \textbf{9.8} \\
\cline{2-12}
& \multirow{2}*{GATE} & Baseline & 12.2 & 15.5 & 16.8 & 7.2 & 9.0 & 9.5 & 5.2 & 7.2 & 8.5 \\
& & TDE & \textbf{18.5} & \textbf{24.9} & \textbf{28.3} & \textbf{11.1} & \textbf{13.9} & \textbf{15.2} & \textbf{6.6} & \textbf{8.5} & \textbf{9.9} \\
\hline
\multirow{4}*{VTransE\textsuperscript{$\dagger$}} & \multirow{2}*{SUM} & Baseline & 11.6 & 14.7 & 15.8 & 6.7 & 8.2 & 8.7 & 3.7 & 5.0 & 6.0 \\
& & TDE & \textbf{17.3} & \textbf{24.6} & \textbf{28.0} & \textbf{9.3} & \textbf{12.9} & \textbf{14.8} & \textbf{6.3} & \textbf{8.6} & \textbf{10.5} \\
\cline{2-12}
& \multirow{2}*{GATE} & Baseline & 13.6 & 17.1 & 18.6 & 6.6 & 8.2 & 8.7 & 5.1 & 6.8 & 8.0 \\
& & TDE & \textbf{18.9} & \textbf{25.3} & \textbf{28.4} & \textbf{9.8} & \textbf{13.1} & \textbf{14.7} & \textbf{6.0} & \textbf{8.5} & \textbf{10.2} \\
\hline
\multirow{4}*{VCTree\textsuperscript{$\dagger$}} & \multirow{2}*{SUM} & Baseline & 11.7 & 14.9 & 16.1 & 6.2 & 7.5 & 7.9 & 4.2 & 5.7 & 6.9 \\
& & TDE & \textbf{18.4} & \textbf{25.4} & \textbf{28.7} & \textbf{8.9} & \textbf{12.2} & \textbf{14.0} & \textbf{6.9} & \textbf{9.3} & \textbf{11.1}\\
\cline{2-12}
& \multirow{2}*{GATE} & Baseline & 12.4 & 15.4 & 16.6 & 6.3 & 7.5 & 8.0 & 4.9 & 6.6 & 7.7\\
& & TDE & \textbf{17.2} & \textbf{23.3} & \textbf{26.6} & \textbf{8.9} & \textbf{11.8} & \textbf{13.4} & \textbf{6.3} & \textbf{8.6} & \textbf{10.3} \\
\hline
\hline
\end{tabular}
}
\caption{The SGG performances of Relationship Retrieval on mean Recall@K~\cite{tang2019learning, chen2019knowledge}. The SGG models re-implemented under our codebase are denoted by the superscript $\dagger$.}
\label{tab:1}
\vspace{-0.2in}
\end{table*}

\vspace{-0.05in}
\subsection{Scene Graph Generation Diagnosis}
\vspace{-0.05in}
\label{sec:metrics}
Our proposed SGG diagnosis has the following three evaluations:

\noindent\textbf{1. Relationship Retrieval (RR).} It can be further divided into three sub-tasks: (1) Predicate Classification \textbf{(PredCls)}: taking ground truth bounding boxes and labels as inputs, (2) Scene Graph Classification \textbf{(SGCls)}: using ground truth bounding boxes without labels, (3) Scene Graph Detection \textbf{(SGDet)}: detecting SGs from scratch. The conventional metric of RR is \textbf{Recall@K (R@K)}, which was abandoned in this paper due to the reporting bias~\cite{misra2016seeing}. As illustrated in Figure~\ref{fig:3}(b), previous methods like \cite{zellers2018neural} with good performance on R@K unfairly cater to ``head'' predicates, \eg, \texttt{on}, while neglect the ``tail'' ones, \eg, predicates like \texttt{parked on}, \texttt{laying on} have embarrassingly 0.0 Recall@100. To speak for the valuable ``tail'' rather than the trivial ``head'', we adopted a recent replacement, \textbf{mean Recall@K (mR@K)}, proposed by Tang~\etal~\cite{tang2019learning} and Chen~\etal~\cite{chen2019knowledge}. mR@K retrieves each predicate separately and then averages R@K for all predicates. 

\noindent\textbf{2. Zero-Shot Relationship Retrieval (ZSRR).} It was introduced by Lu~\etal~\cite{lu2016visual} as \textbf{Zero-Shot Recall@K} and was firstly evaluated on VG dataset in this paper, which only reports the R@K of those subject-predicate-object triplets that have never been observed in the training set. ZSRR also has three sub-tasks as RR.

\noindent\textbf{3. Sentence-to-Graph Retrieval (S2GR).} It uses the image caption sentence as the query to retrieve images represented as SGs. Both RR and ZSRR are triplet-level evaluations, ignoring the graph-level coherence. Therefore, we design S2GR, using human descriptions to retrieve detected SGs. We didn't use proxy vision-language tasks like captioning~\cite{yang2019auto, yao2018exploring} and VQA~\cite{teney2017graph, hudson2019gqa} as the diagnosis, because their implementations have too many components unrelated to SGG and their datasets are challenged by their own biases~\cite{agrawal2018don, hendricks2018women, manjunatha2019explicit}. In S2GR, the detected SGs (using SGDet) are regarded as the only representations of images, cut off all the dependencies on black-box visual features, so any bias on SGG would sensitively violate the coherence of SGs, resulting in worse retrieval results. For example, if \texttt{walking on} was detected as the biased alternative \texttt{on}, images would be mixed up with those have \texttt{sitting on} or \texttt{laying on}. Note that S2GR is fundamentally different from the previous image retrieval with scene graph~\cite{Johnson_2015_CVPR, schuster2015generating}, because the latter still consider the images as visual features but not SGs. \textbf{Recall@20/100 (R@20/100)} and median ranking indexes of retrieved results \textbf{(Med)} on the gallery size of 1,000 and 5,000 were evaluated. Note that S2GR should have diverse implementations as long as its spirit: graph-level symbolic retrieval, is fulfilled. We provide our implementation in the next sub-section.

\begin{table}
\centering
\scalebox{0.7}
{
\begin{tabular}{r | r | r | c | c | c}
\hline
\multicolumn{3}{c}{Zero-Shot Relationship Retrieval} &\multicolumn{1}{c}{PredCls} & \multicolumn{1}{c}{SGCls} & \multicolumn{1}{c}{SGDet} \\
\hline
Model & Fusion & Method & R@50/100 & R@50/100 & R@50/100 \\ 
\hline 
\multirow{12}*{MOTIFS\textsuperscript{$\dagger$}} & \multirow{10}*{SUM} & Baseline & 10.9~/~14.5 & 2.2~/~3.0 & 0.1~/~0.2 \\
\cline{3-6}
& & Focal & 10.9~/~14.4 & 2.2~/~3.1 & 0.1~/~0.3 \\
& & Reweight & 0.7~/~0.9 & 0.1~/0.1 & 0.0~/~0.0 \\
& & Resample  & 11.1~/~14.3 & 2.3~/~3.1 & 0.1~/~0.3\\
\cline{3-6}
& & X2Y  & 11.8~/~17.6 & 2.3~/~3.7 & 1.6~/~2.7\\
& & X2Y-Tr  & 13.7~/~17.6 & 3.1~/~4.2 & 1.8~/~2.8\\
\cline{3-6}
& & TE  & 14.2~/~18.1 & 1.4~/~2.0 & 1.4~/~1.8\\
& & NIE  & 2.4~/~3.2 & 0.2~/~0.4 & 0.3~/~0.6\\
& & TDE & \textbf{14.4~/~18.2} & \textbf{3.4~/~4.5} & \textbf{2.3~/~2.9} \\
\cline{2-6}
& \multirow{2}*{GATE} & Baseline & 7.4~/~10.6 & 0.9~/~1.3 & 0.2~/~0.4 \\
& & TDE & \textbf{7.7~/~11.0} & \textbf{1.9~/~2.6} & \textbf{1.9~/~2.5} \\
\hline 
\multirow{4}*{VTransE\textsuperscript{$\dagger$}} & \multirow{2}*{SUM} & Baseline & 11.3~/~14.7 & 2.5~/~3.3 & 0.8~/~1.5  \\
& & TDE & \textbf{13.3~/~17.6} & \textbf{2.9~/~3.8} & \textbf{2.0~/~2.7} \\
\cline{2-6}
& \multirow{2}*{GATE} & Baseline & 4.2~/~5.9 & 1.9~/~2.6 & \textbf{1.9}~/~2.6 \\
& & TDE & \textbf{5.3~/~7.9} & \textbf{2.1~/~3.0} & \textbf{1.9~/~2.7} \\
\hline 
\multirow{4}*{VCTree\textsuperscript{$\dagger$}} & \multirow{2}*{SUM} & Baseline & 10.8~/~14.3 & 1.9~/~2.6 & 0.2~/~0.7 \\
& & TDE & \textbf{14.3~/~17.6} & \textbf{3.2~/~4.0} & \textbf{2.6~/~3.2} \\
\cline{2-6}
& \multirow{2}*{GATE} & Baseline & 4.4~/~6.8 & 2.5~/~3.3 & 1.8~/~2.7\\
& & TDE & \textbf{5.9~/~8.1} & \textbf{3.0~/~3.7} & \textbf{2.2~/~2.8} \\
\hline
\hline
\end{tabular}
}
\caption{The results of Zero-Shot Relationship Retrieval.}
\label{tab:2}
\vspace{-0.2in}
\end{table}

\vspace{-0.05in}
\subsection{Implementation Details}
\vspace{-0.05in}
\noindent\textbf{Object Detector.} Following the previous works~\cite{xu2017scene, zellers2018neural, tang2019learning}, we pre-trained a Faster R-CNN~\cite{ren2015faster} and froze it to be the underlying detector of our SGG models. We equipped the Faster R-CNN with a ResNeXt-101-FPN~\cite{lin2017feature, xie2017aggregated} backbone and scaled the longer side of input images to be 1k pixels. The detector was trained on the training set of VG using SGD as optimizer. We set the batch size to 8 and the initial learning rate to $8\times 10^{-3}$, which was decayed by the factor of 10 on the 30k\textsuperscript{th} and 40k\textsuperscript{th} iterations. The final detector achieved $28.14$ mAP on VG test set (using 0.5 IoU threshold). 4 2080ti GPUs were used for the pre-training. 

\begin{table}
\centering
\scalebox{0.6}
{
\begin{tabular}{r | r | r | c c | c | c c | c }
\hline
\multicolumn{9}{c}{Sentence-to-Graph Retrieval}\\
\hline
\multicolumn{2}{r}{Gallery Size} & \multicolumn{1}{r}{} & \multicolumn{3}{|c}{1000} & \multicolumn{3}{|c}{5000} \\
\hline
Model & Fusion & Method & R@20 & R@100 & Med & R@20 & R@100 & Med \\
\hline 
\multirow{12}*{MOTIFS\textsuperscript{$\dagger$}} & \multirow{10}*{SUM} & Baseline & 11.6 & 39.9 & 155 & 3.1 & 12.1 & 708\\
\cline{3-9}
& & Focal & 10.9 & 39.0 & 163 & 2.9 & 11.1 & 737 \\
& & Reweight & 9.7 & 36.8 & 159 & 3.0 & 11.4 & 725 \\
& & Resample  & 13.1 & 43.6 & 124 & 2.5 & 13.4 & 593 \\
\cline{3-9}
& & X2Y  & 14.3 & 44.8 & 125 & 3.5 & 14.6 & 556 \\
& & X2Y-Tr  & 14.5 & 45.6 & 114 & 3.9 & 16.8 & 525 \\
\cline{3-9}
& & TE  & 15.9 & 49.9 & 100 & 4.4 & 16.9 & 469 \\
& & NIE  & 6.7 & 29.2 & 202 & 1.6 & 8.6 & 1050 \\
& & TDE & \textbf{17.0} & \textbf{53.6} & \textbf{91} & \textbf{5.2} & \textbf{18.9} & \textbf{425} \\
\cline{2-9}
& \multirow{2}*{GATE} & Baseline & 13.7 & 45.6 & 143 & 4.4 & 16.2 & 618 \\
& & TDE & \textbf{20.8} & \textbf{59.2} & \textbf{72} & \textbf{5.2} & \textbf{21.3} & \textbf{325} \\
\hline 
\multirow{4}*{VTransE\textsuperscript{$\dagger$}} & \multirow{2}*{SUM} & Baseline & 12.3 & 42.3 & 129 & \textbf{3.6} & 15.0 & 596 \\
& & TDE & \textbf{14.7} & \textbf{48.4} & \textbf{106} & \textbf{3.6} & \textbf{16.3} & \textbf{483} \\
\cline{2-9}
& \multirow{2}*{GATE} & Baseline & 12.9 & 41.8 & 136 & 3.8 & 14.3 & 634 \\
& & TDE & \textbf{18.5} & \textbf{50.4} & \textbf{110} & \textbf{4.5} & \textbf{19.1} & \textbf{486} \\
\hline 
\multirow{4}*{VCTree\textsuperscript{$\dagger$}} & \multirow{2}*{SUM} & Baseline & 9.9 & 37.4 & 150 & 3.1 & 11.5 & 745 \\
& & TDE & \textbf{19.0} & \textbf{57.0} & \textbf{82} & \textbf{5.0} & \textbf{20.0} & \textbf{385} \\
\cline{2-9}
& \multirow{2}*{GATE} & Baseline & 13.4 & 44.1 & 121 & 3.7 & 13.6 & 583 \\
& & TDE & \textbf{19.1} & \textbf{55.5} & \textbf{87} & \textbf{5.1} & \textbf{20.3} & \textbf{395} \\
\hline
\hline
\end{tabular}
}
\caption{The results of Sentence-to-Graph Retrieval.}
\label{tab:3}
\vspace{-0.2in}
\end{table}

\noindent\textbf{Scene Graph Generation.} On top of the frozen detector, we trained SGG models using SGD as optimizer. Batch size and initial learning rate were set to be 12 and $12\times 10^{-2}$ for PredCls and SGCls; 8 and $8\times 10^{-2}$ for SGDet. The learning rate would be decayed by 10 two times after the validation performance plateaus. For SGDet, 80 RoIs were sampled for each image and Per-Class NMS~\cite{rosenfeld1971edge, zellers2018neural} with 0.5 IoU was applied in object prediction. We sampled up to 1,024 subject-object pairs containing 75\% background pairs during training. Different from previous works~\cite{zellers2018neural, tang2019learning, chen2019counterfactual}, we didn't assume that non-overlapping subject-object pairs are invalid in SGDet, making SGG more general.

\noindent\textbf{Sentence-to-Graph Retrieval.} We handled S2GR as a graph-to-graph matching problem. The query captions of each image were stuck together and parsed to a text-SG using \cite{schuster2015generating}. We set all the subject/object and predicates that appear less than 5 times to ``UNKNOWN'' tokens, obtaining a dictionary of size 4,459 subject/object entities and 645 predicates, respectively. The original image SG generated from SGDet contains a fixed number of RoIs and forces all valid subject-object pairs to predict foreground relationships, to serve the $K$ number in mR@K, which is inappropriate for S2GR. Therefore, we used a threshold of 0.1 to filter RoIs by the label probabilities and removed all background predicates from the graph. Recall that the vocabulary size of the entity and predicate for image SGs are 150 and 50 as we mentioned before. To match the two heterogeneous graphs: image SG and text SG, in a unified space, we used BAN~\cite{kim2018bilinear} to encode the two graph types into fixed-dimension vectors to facilitate the retrieval. More details can be found in supplementary material.

\begin{figure}[t]
   \begin{minipage}[b]{1.0\linewidth}
   \centerline{\includegraphics[width=85mm]{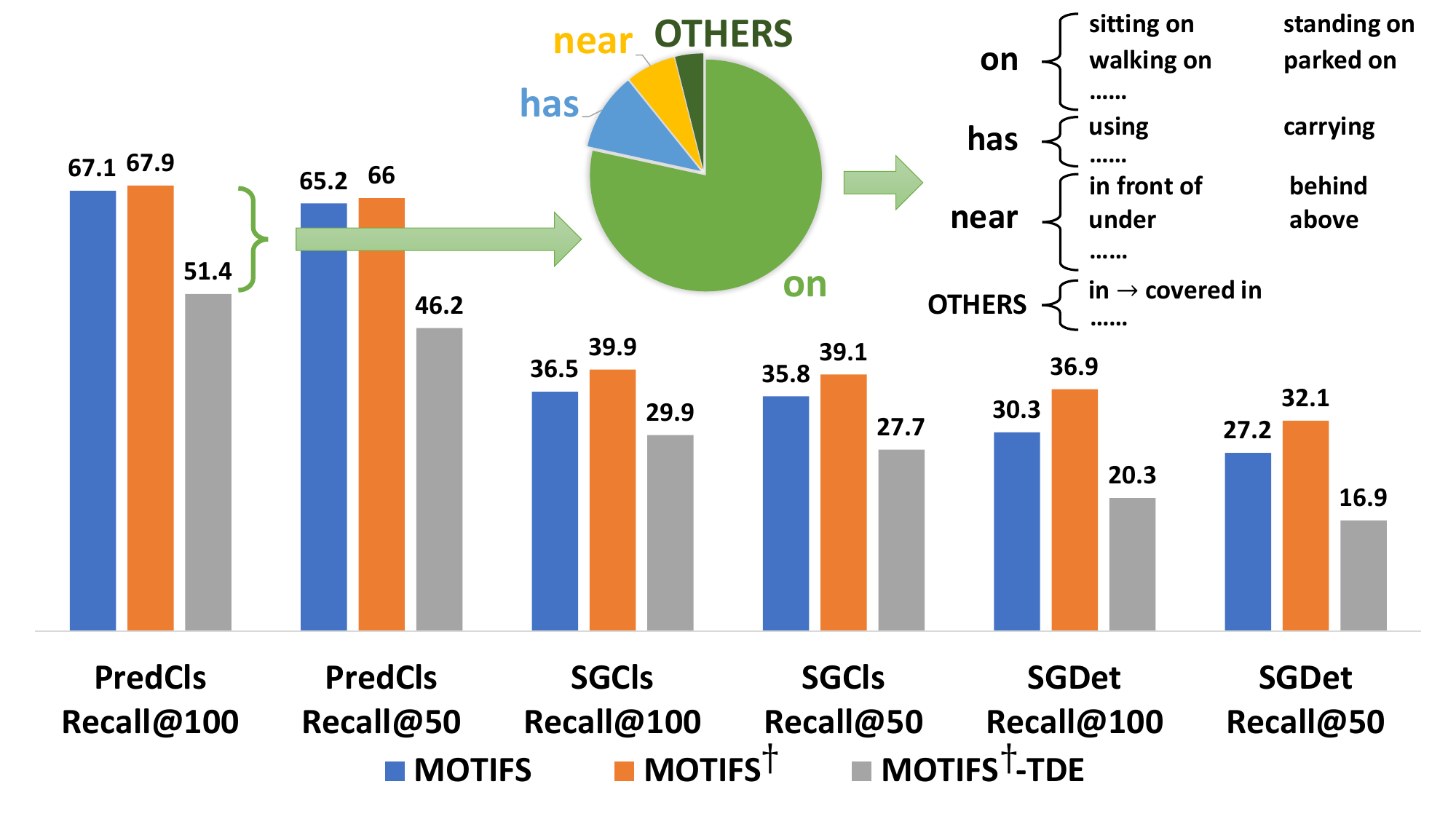}}
   \end{minipage}
   \caption{The pie chart summarizes all the relationships, that are correctly detected by the baseline model but considered ``incorrect'' by TDE.  The right side of the pie chart shows the corresponding labels given by the TDE. Combining with our qualitative examples, we believe that the drop of Recall@K is caused by two reasons: 1) the annotators’ preference towards simple annotations caused by bounded rationality~\cite{simon1990bounded}, 2) TDE tends to predict more action-like relationships rather than vague prepositions.}
   \label{fig:5} 
   \vspace{-0.2in}
\end{figure}

\begin{figure*}
   \begin{minipage}[b]{1.0\linewidth}
   \centerline{\includegraphics[width=170mm]{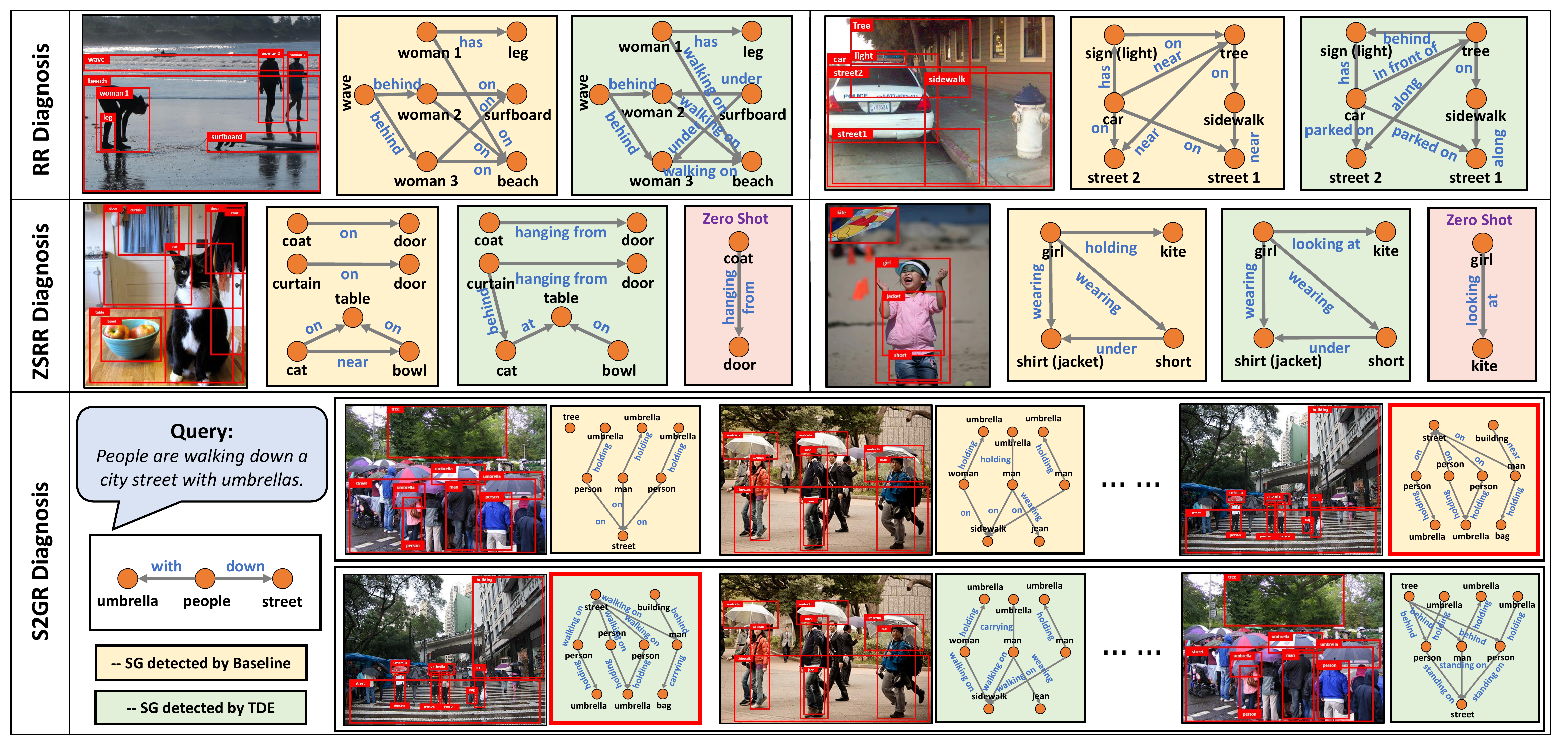}}
   \end{minipage}
   \caption{ Results of scene graphs generated from MOTIF\textsuperscript{$\dagger$}-SUM baseline (yellow) and corresponding TDE (green). Top: relationship retrieval results. Mid: zero shot relationship retrieval results. Red boxes indicate the zero shot triplets. Bottom: results of S2GR. Red boxes mean the correctly retrieved SGs. Part of the trivial detected objects are removed from the graphs, due to space limitation.}
   \label{fig:7} 
   \vspace{-0.2in}
\end{figure*}

\vspace{-0.05in}
\subsection{Ablation Studies}
\vspace{-0.05in}
Except for the models and fusion functions that we've discussed before, we also investigated three conventional debiasing methods, two intuitive causal graph surgeries, and other two types of causal effects: 1) \textbf{Focal}: focal loss~\cite{lin2017focal} automatically penalizes well-learned samples and focuses on the hard ones. We followed the  hyper-parameters ($\gamma=2.0, \alpha=0.25$) optimized in~\cite{lin2017focal}. 2) \textbf{Reweight}: weighted cross-entropy is widely used in the industry for biased data. The inversed sample fractions were assigned to each predicate category as weights. 3) \textbf{Resample}~\cite{burnaev2015influence}: rare categories were up-sampled by the inversed sample fraction during training. 4) \textbf{X2Y}: since we argued that the unbiased effect was under the effect of object features $X$, it directly generated SG by the outputs of $X\to Y$ branch after biased training. 5) \textbf{X2Y-Tr}: it even cut off other branches, using $X\to Y$ for both training and testing. 6) \textbf{TE}: as we introduced in Section~\ref{sec:unbiased_pred}, TE is the debiasing method that not conditioned on the contexts. 7) \textbf{NIE}: it is the marginal difference between TDE and TE, \ie, NIE = TE-TDE, which can be considered as the pure effect caused by introducing the bias $Z\to Y$. \textbf{NOTE}: although zero vector can also be used as the wiped-out input $\bar{x}$, we chose the mean feature of training set for minor improvements.



\vspace{-0.05in}
\subsection{Quantitative Studies}
\vspace{-0.05in}
\noindent\textbf{RR \& ZSRR.} The results are listed in Table~\ref{tab:1}\&~\ref{tab:2}. Despite the fact that conventional debiasing methods: Reweight and Resample, directly hack the mR@K metric, they only gained limited advantages in RR but not in ZSRR. In contrast to the high mR@K of Reweight in RR SGDet, it got embarrassingly $0.0/0.0$ in ZSRR SGDet, indicating that such debiased training methods ruin the useful context prior. Focal loss~\cite{lin2017focal} barely worked for both RR and ZSRR. Causal graph surgeries, X2Y and X2Y-Tr, both improved RR and ZSRR from the baseline, yet their increases were limited. TE had a very similar performance to TDE, but as we discussed, it removed the general bias rather than the subject-object specific bias. NIE is the marginal improvements from TE to TDE, which was even worse than baseline. Although R@K is not a qualified metric for RR as we discussed, we still reported the R@50/100 performance of MOTIFS\textsuperscript{$\dagger$}-SUM in Figure~\ref{fig:5}. We can observe a performance drop from baseline to TDE, but a further analysis shows that those considered as correct in baseline and ``incorrect'' in TDE were mainly the ``head'' predicates, and they are classified by TDE into more fine-grained ``tail'' classes. Among all three models and two fusion functions, even the worst TDE performance outperforms previous state-of-the-art methods~\cite{tang2019learning, chen2019knowledge} by a large margin on RR mR@K.

\noindent\textbf{S2GR.} In S2GR, Focal and Reweight are even worse than the baseline. Among all the three conventional debiasing methods, Resample was the most stable one based on our experiments. X2Y and X2Y-Tr have minor advantages over baseline. TE takes the 2nd place and was only a little bit worse than TDE. NIE is the worst as we expected because it is only based on the pure context bias. It is worth highlighting that all the three models and two fusion functions had significant improvements after we applied TDE.

\subsection{Qualitative Studies}
We visualized several SGCls examples that generated from MOTIFS\textsuperscript{$\dagger$}-SUM baseline and TDE in the top and mid rows of Figure~\ref{fig:7}, scene graphs generated by TDE are much more discriminative compared to the baseline model which prefers trivial predicates like \texttt{on}.  The right half of the mid row shows that the baseline model would even generate \texttt{holding} due to the long-tail bias when the girl is not touching the kite, implying that the biased predictions are easy to be ``blind'', while TDE successfully predicted \texttt{looking at}. The bottom of Figure~\ref{fig:7} is an example of S2GR, where the SGs detected by baseline model lost the detailed actions of people, considering both \texttt{person walking on street} and \texttt{person standing on street} as \texttt{person on street}, which caused worse retrieval results. All the examples show a clear trend that TDE is much more sensitive to those semantically informative relationships instead of the trivially biased ones.

\section{Conclusions}
We presented a general framework for unbiased SGG from biased training, and this is the first work addressing the serious bias issue in SGG. With the power of \emph{counterfactual causality}, we can remove the harmful bias from the good context bias, which cannot be easily identified by traditional debiasing methods such as data augmentation~\cite{geirhos2018imagenettrained, he2009learning} and unbiased learning~\cite{lin2017focal}. We achieved the unbiasedness by calculating the Total Direct Effect (TDE) with the help of a causal graph, which is a roadmap for training any SGG model. By using the proposed Scene Graph Diagnosis toolkit, our unbiased SGG results are considerably better than their biased counterparts.\\
\noindent\textbf{Acknowledgments~}
We'd like to thank all reviewers for their constructive comments. This work was partially supported by the NTU-Alibaba JRI.


{\small
\bibliographystyle{ieee}
\bibliography{egbib}
}

\clearpage
\appendix
\begin{abstract}
This supplementary document is organized as follows: 1) section~\ref{sec:review}: a comprehensive review of causal effect analysis in causal inference; 2) section~\ref{sec:networks}: more details of the simplified network structures in the original paper; 3) section~\ref{sec:quantitative}: more quantitative studies; 4) section~\ref{sec:qualitative}: more qualitative studies.
\end{abstract}

\section{Review of Causal Effect Analysis}
\label{sec:review}
In this section, a comprehensive review of causal effect analysis is given in the form of the causal graph we proposed in Section 3, and we still follow the notations from the original paper. More detailed background knowledge about causal inference can be found in \cite{pearl2016causal, Judea2018thebookofwhy} while the extension of effect analysis (a.k.a. mediation analysis) is given in \cite{robins1992identifiability, pearl2001direct, vanderweele2013three, vanderweele2015explanation}. 

\subsection{Mediator}
Since the exhaustive introduction of causal inference would beyond the scope of this paper, we simplified or skipped the definitions of several concepts in the original paper without affecting the understanding. One of the skipped concepts is the mediator. In a causal graph, when we care about the effect of a variable $X$ to the output variable $Y$, the descendant node of $X$ that is located in the path between them is the mediator. For example, in the study of carcinogenesis by smoke ($\text{Cigarette}\to \text{Nicotine}\to \text{Cancer}$), nicotine is the mediator. In our case, object labels $Z$ is the mediator of $X$ to $Y$, which can be considered as the side effect of $X$ that also affects $Y$. 

\subsection{Total, Direct and Indirect Effects}
\begin{figure}[t!]
   \begin{minipage}[b]{1\linewidth}
   \centerline{\includegraphics[width=80mm]{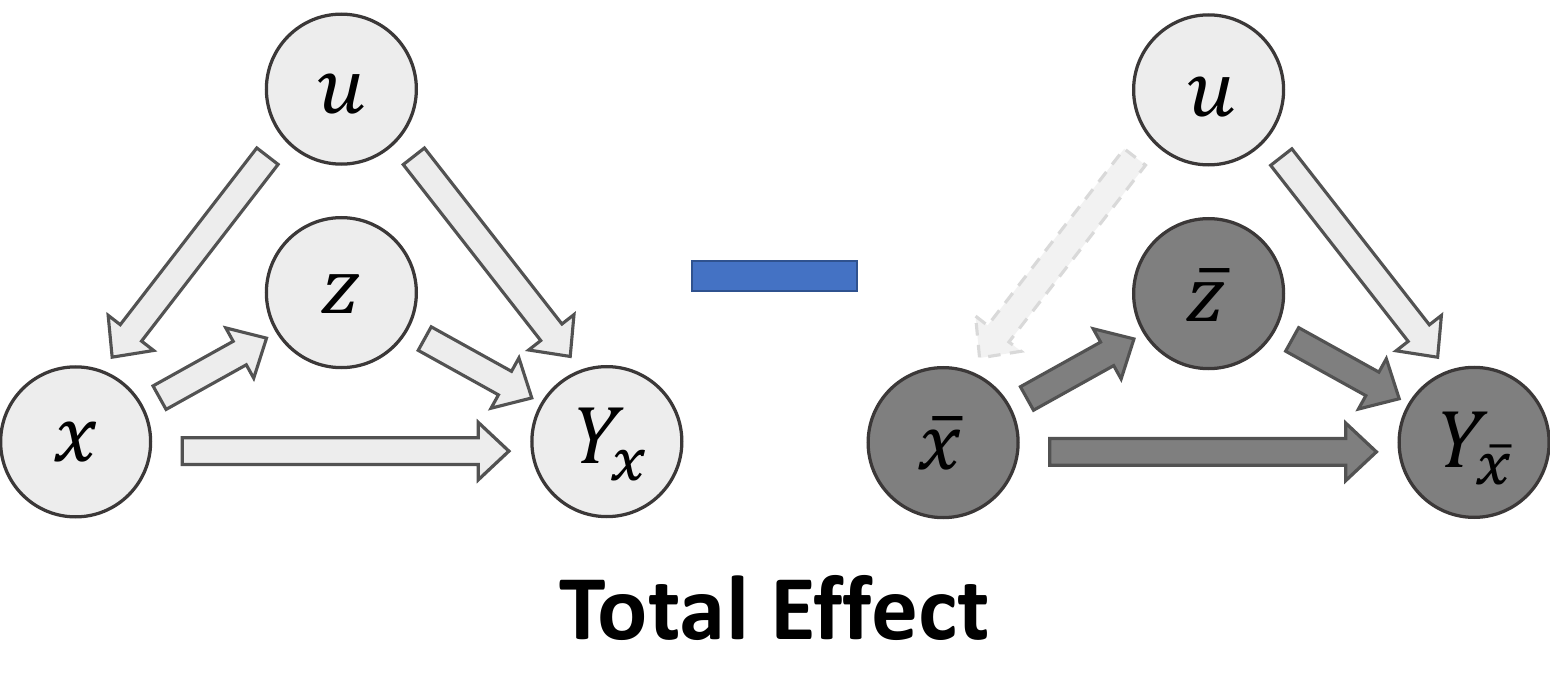}}
   \end{minipage}
   \caption{The illustration of Total Effect on causal graph.}
   \label{supp_fig:te} 
\end{figure}

As we discussed in Section 4.2, without further counterfactual intervention on the mediator $Z$, the overall effect of $X$ towards $Y$ is regarded as the Total Effect (TE) of $X$ on $Y$, which can be calculated as:
\begin{equation}
    TE=Y_x(u) - Y_{\bar{x}}(u). 
\end{equation}
As illustrated in Figure~\ref{supp_fig:te}, other than the path $I\to X$ that is cut off by the intervention $X=\bar{x}$, all the other variables will take their values through the links of causal graph. Especially, the mediator $Z$ will get value $\bar{z}$, which is calculated from Eq. (2) given $\bar{x}$ as input.

However, by only using the TE, we are still not able to separate the mediator-specific ``causal effect'' from ``side effect'', which limits the value of causal effect analysis. Thanks to the development of causal inference, here comes the decomposition of TE~\cite{pearl2001direct, vanderweele2013three}. Generally, the TE of $X$ is composed of the Direct Effect (DE) caused by the causal path $X\to Y$ and Indirect Effect (IE) caused by the side-effect path $X\to Z\to Y$. Depending on whose effect we want to obtain, two kinds of decomposition can be applied.

\begin{figure}[t!]
   \begin{minipage}[b]{1\linewidth}
   \centerline{\includegraphics[width=80mm]{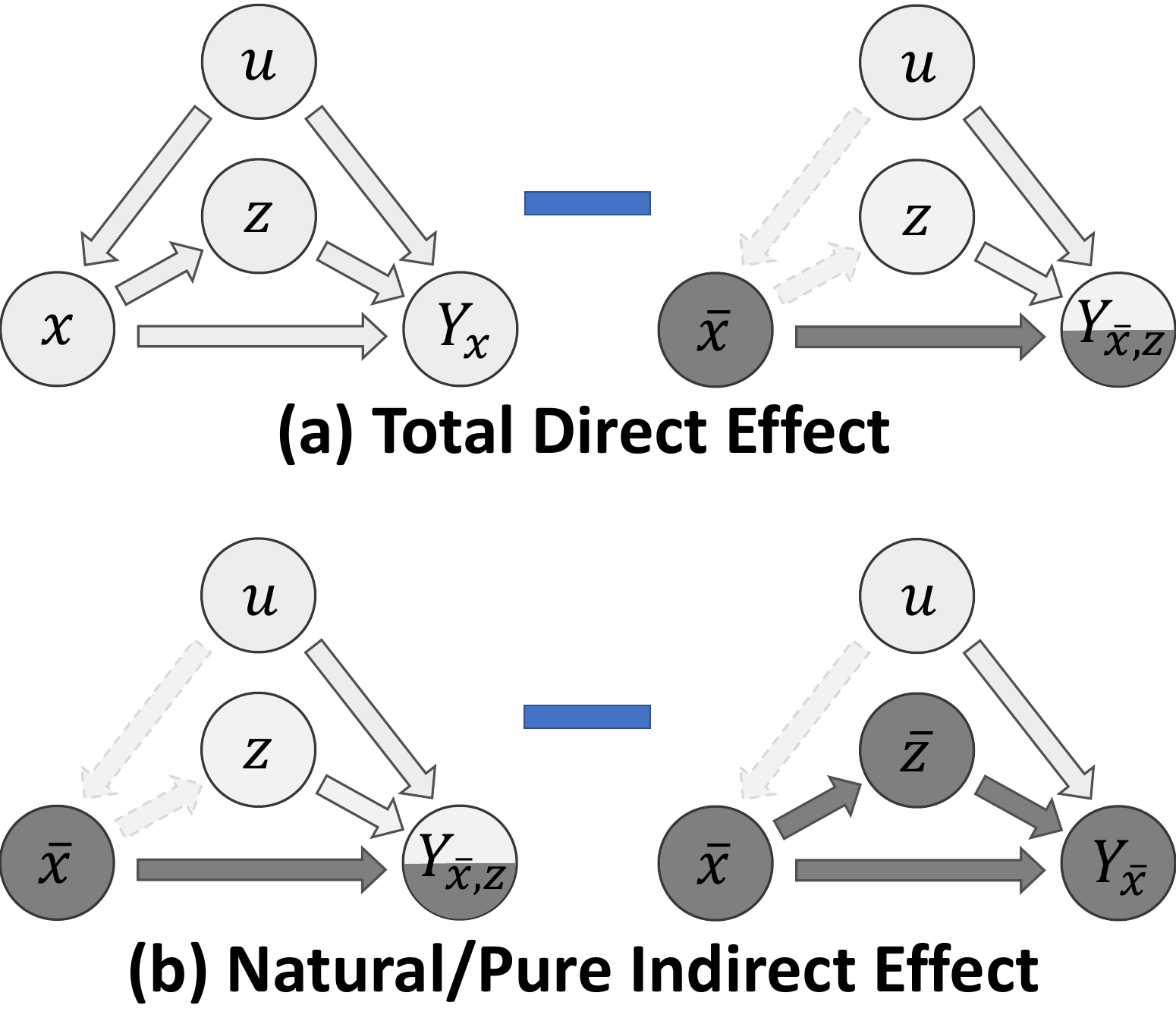}}
   \end{minipage}
   \caption{The illustration of Total Direct Effect and Pure/Natural Indirect Effect on causal graph.}
   \label{supp_fig:tde_nie} 
\end{figure}

\noindent\textbf{Decomposition 1:} The first kind of decomposition is what we used in the Section 4.2, which separates the TE into the Total Direct Effect (TDE) and the Natural/Pure Indirect Effect (NIE/PIE). The former one has already been defined in the original paper as:
\begin{equation}
    TDE=Y_x(u)-Y_{\bar{x}, z}(u),
\end{equation}
which can be regarded as the effect of $X$ in the real situation, \ie, $Z$ always takes the value $z$ as if it had seen the real $x$. Meanwhile, the NIE or PIE is the effect caused by the mediator $Z$ under a pure/natural situation, \ie, $X$ will not take the value $x$ under the specific case and it's only assigned to the general unactivated value $\bar{x}$. Therefore, the NIE of $Z$ is denoted as:
\begin{align}
    NIE &=Y_{\bar{x}, z}(u) - Y_{\bar{x}}(u) \\
        &=TE - TDE,
\end{align}
where we can easily identify that NIE is the effect of $Z$ when it changes from $\bar{z}$ to $z$ in a pure environment, \ie, $X=\bar{x}$. The illustrations of TDE and NIE are given in Figure~\ref{supp_fig:tde_nie}. 

\noindent\textbf{Decomposition 2:} The second type of decomposition is opposite to the first one. It's mainly adopted when the indirect effect of the mediator is what we are looking for. For example, in the study of carcinogenesis by smoke ($\text{Cigarette}\to \text{Nicotine}\to \text{Cancer}$), sometimes the side effect of Nicotine is what researchers really care about. In this case, TE can be decomposed into Total Indirect Effect(TIE) and Natural/Pure Direct Effect (NDE/PDE). The definition of the former one is very similar to the NIE except for the environment being the real case $X=x$, which is therefore formulated as:
\begin{equation}
    TIE = Y_x(u)-Y_{x, \bar{z}}(u).
\end{equation}
At the same time, since direct effect is not the target, their pure/natural effect should be removed from the TE. The calculation of NDE/PDE is following:
\begin{align}
    NDE &= Y_{x, \bar{z}}(u) - Y_{\bar{x}}(u) \\
        &= TE - TIE,
\end{align}
where NDE is the effect of $X$ changing from $\bar{x}$ to $x$ under the pure environment $Z=\bar{z}$. In general, we should put the effect we care under the real environment, \ie TDE or TIE, so we can get the results specific to each cases. 

The above two types of decomposition are both commonly used in medical, political or psychological research~\cite{richiardi2013mediation, keele2015statistics, dunn2015evaluation, mackinnon2007mediation, king2008political}, which depends on which effect we want to obtain, main effect or side effect. Note that, if the system is a pure linear system, both two types of decomposition would be exactly the same. 

\begin{figure}[t!]
   \begin{minipage}[b]{1\linewidth}
   \centerline{\includegraphics[width=80mm]{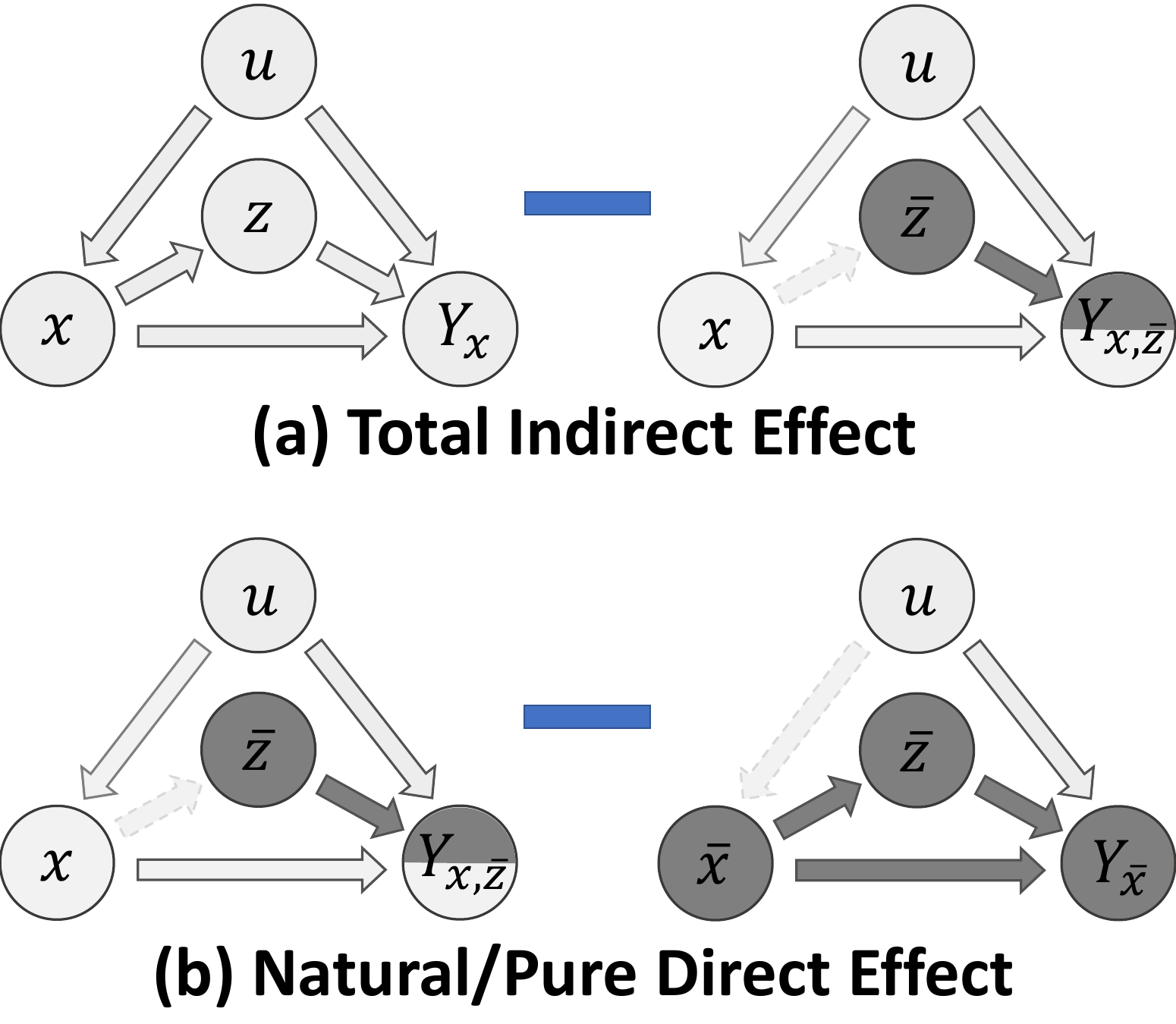}}
   \end{minipage}
   \caption{The illustration of Total Indirect Effect and Pure/Natural Direct Effect on causal graph.}
   \label{supp_fig:tie_nde} 
\end{figure}

\section{Network Details}
\label{sec:networks}

\subsection{Scene Graph Generation}
In the original paper, we simplified the feature extraction module in Link $I\to X$ and the visual context module in Link $I\to Y$. Their details will be given in this subsection.

\noindent\textbf{Feature Extraction Module.} Since we adopted ResNeXt-101-FPN~\cite{lin2017feature, xie2017aggregated} as the backbone, the extracted $\mathcal{M}$ contains feature maps from 4 scales: $(1/4, 1/8, 1/16, 1/32) \to (\mathcal{M}_0, \mathcal{M}_1, \mathcal{M}_2, \mathcal{M}_3)$. Each bounding box will be assigned to the corresponding $\mathcal{M}_k, (k=0,1,2,3)$ based on their areas~\cite{massa2018mrcnn}. Given a bounding box $b_i$ with area $a_i$, the corresponding index $k$ of feature map is calculated as follows:
\begin{equation}
    k = \max(2, \min(5,\floor*{4+\log_2(a_i / 224 + 1\times 10^{-6})})) - 2.
\end{equation}
Then ROIAlign~\cite{he2017mask} will be applied to the selected bounding box $b_i$ on the corresponding $\mathcal{M}_k$ for the feature $r_i$ as we described in Section 3.

\noindent\textbf{Visual Context Module.} To extract the visual context feature $v'_e$ for the union box $b_i \cup b_j$, we consider all 4 feature maps will provide complementary contextual information from different levels. Therefore, we extract ROIAlign~\cite{he2017mask} features on all 4 feature maps before we project the visual context feature into a feature space of $\mathbb{R}^{4096}$. The entire module is summarized in the Table~\ref{supp_tab:1}, where the dummy mask operation in (7) generates two masks for $b_i$ and $b_j$ independently, assigning 1.0 to the pixels inside the bounding box and 0.0 for the rest.

\begin{table}
\centering
\scalebox{0.8}
{
\begin{tabular}{| c | c | c | c |}
\hline
Index & Input & Operation & Output \\
\hline 
(1) & ($\mathcal{M}_0, b_i \cup b_j$) & ROIAlign & ($7\times 7\times 256$) \\
\hline 
(2) & ($\mathcal{M}_1, b_i \cup b_j$) & ROIAlign & ($7\times 7\times 256$) \\
\hline 
(3) & ($\mathcal{M}_2, b_i \cup b_j$) & ROIAlign & ($7\times 7\times 256$) \\
\hline 
(4) & ($\mathcal{M}_3, b_i \cup b_j$) & ROIAlign & ($7\times 7\times 256$) \\
\hline
(5) & (1-4) & Concatenation & ($7\times 7\times 1024$) \\
\hline
(6) & (5) & Conv & ($7\times 7\times 256$) \\
\hline
(7) & $b_i, b_j$ & dummy mask & ($27\times 27\times 2$) \\
\hline
(8) & (7) & Conv+ReLU+BatchNorm & ($14\times 14\times 128$) \\
\hline
(9) & (8) & MaxPool & ($7\times 7\times 128$) \\
\hline
(10) & (9) & Conv+Relu+BatchNorm & ($7\times 7\times 256$) \\
\hline
(11) & (6),(10) & Element-wise Addition & ($7\times 7\times 256$) \\
\hline
(12) & (11) & Flatten & 12,544 \\
\hline
(13) & (12) & FC+ReLU & 4,096 \\
\hline
(14) & (13) & FC+ReLU & 4,096 \\
\hline
\end{tabular}
}
\caption{The details of Visual Context Module.}
\label{supp_tab:1}
\end{table}

\noindent\textbf{The Special Treatment for PredCls.} In the original paper, we skipped a special case of causal graph, \ie, causal graph for Predicate Classification (PredCls), for simplification. In PredCls, the ground truth object labels are given, which means the link $X\to Z$ is blocked by assigning ground truth labels. It won't affect TDE calculation, where $Z$ takes the real value $z$. However, it's involved in the ablation studies of TE and NIE, where $Z$ could be assigned to $\bar{z}$. In this case, $\bar{z}$ will directly use to the mean vector of training set rather than be calculated from Eq.(2). We also need to notice that, for MOTIFS~\cite{zellers2018neural}, Eq.(3) will take $z_e$ as input too, which is simplified in the original paper, because $z_e$ itself is derived from $x_e$ and it can be considered as the interaction between link $X\to Y$ and $Z\to Y$ in the causal graph.

\subsection{Sentence-to-Graph Retrieval}
As we mentioned in the original paper, we treated Sentence-to-Graph Retrieval (S2GR) as the graph-to-graph matching problem, parsing query captions to text-SGs by \cite{schuster2015generating}. Both detected image-SGs and parsed text-SGs are composed of entities $E^k=\{e^k_i\}$ and relationships $R^k=\{r^k_{ij}=(s^k_i,p^k_{ij},o^k_j)\}$, where $k\in\{text, image\}$, subject and object categories ($s^k_i,o^k_j$) share the same dictionary with $e^k_i$ for each $k$, $p^k_{ij}$ denotes the onehot vector of the predicate category. 

The image-SGs and text-SGs are equipped with different embedding layers, because they have different dictionaries. The entities and relationships are encoded as:
\begin{align}
    E_{embed}^k &= W_e^k E^k, \\ 
    R_{embed}^k &= [W_s^k S^k; W_p^k P^k; W_o^k O^k],
\end{align}
where $E_{embed}^k\in \mathbb{R}^{N_d \times N_e^k}$, $R_{embed}^k \in \mathbb{R}^{3N_d \times N_r^k}$, $N_d=512$ is the dimension of embedded feature, $N_e^k, N_r^k$ are numbers of entities and relationships for each image. 

\subsubsection{Bilinear Attention Scene Graph Encoding}
Since entities and relationships are both important for SGs, we apply Bilinear Attention Network (BAN)~\cite{kim2018bilinear} to encode their multimodal interactions into the same representation space. The same BAN model is used for both text-SGs and image-SGs, hence we remove $k$ hereinafter for simplification. The original BAN involves two steps: 1) attention map generation, and 2) bilinear attended feature calculation. Because scene graph has already provides connections between entities and relationships, we skipped the first step and used normalized scene graph connection as attention map $A_{ij}=M_{ij} / \sum_j M_{ij}$, where $A,M\in \mathbb{R}^{N_e \times N_r}$, the scene graph connection $M$ is defined as follows:
\begin{equation}
    M_{ij}= \left\{
    \begin{aligned}
        1 &,\ if\ E_i\ in\ R_j,\\
        0 &,\ if\ E_i\ not\ in\ R_j.
    \end{aligned}
    \right.
\end{equation}
The bilinear attended scene graph encoding is calculated by Table~\ref{supp_tab:3}, where steps (4-10) are calculated 2 times, and the final output $E_{graph}\in \mathbb{R}^{1024}$ is a feature vector representing the whole SG. The same BAN is used for both text-SG or image-SG, \ie, the parameters of the BAN are shared. 

The model was trained by the triplet loss~\cite{schroff2015facenet} with L1 distance. The model was trained in 30 epochs by SGD optimizer and set batch size to be $12$. Learning rate was set to be $12\times 10^{-2}$, which was decayed at 10\textsuperscript{th} and 25\textsuperscript{th} epochs by the factor of 10.

\begin{table}
\centering
\scalebox{0.65}
{
\begin{tabular}{| c | c | c | c | c |}
\hline
Index & Input & Loop & Operation & Output \\
\hline 
(1) & $E_{embed}$ &  & Input Shape & ($N_e\times 512$) \\
\hline 
(2) & $R_{embed}$ &  & Input Shape & ($N_r\times 512$) \\
\hline 
(3) & $A$ & & Input Shape & ($N_e\times N_r$) \\
\hline
(4) & (1) & start & Transpose + Unsqueeze & ($512\times 1\times N_e$) \\
\hline
(5) & (2) & $\downarrow$ & Transpose + Unsqueeze & ($512\times N_r\times 1$) \\
\hline
(6) & (3) & $\downarrow$ & Unsqueeze &  ($1\times N_e\times N_r$) \\
\hline
(7) & (4),(6) & $\downarrow$ & Matrix Multiplication & ($512\times 1\times N_r$) \\
\hline
(8) & (5),(7) & $\downarrow$ & Matrix Multiplication & ($512\times 1\times 1$) \\
\hline
(9) & (8) & $\downarrow$ & Squeeze + FC & ($512$) \\
\hline
(10) & (4),(9) & end & Unsqueeze + Element-wise Addition & ($512\times 1\times N_e$) \\
\hline
(11) & (10) &  & Sum Over $N_e$ & 512 \\
\hline
(12) & (11) & & FC + ReLU + FC + ReLU & 1024 \\
\hline
\end{tabular}
}
\caption{The details of Bilinear Attention Scene Graph Encoding Module.}
\label{supp_tab:3}
\end{table}

\section{Quantitative Studies}
\label{sec:quantitative}

\begin{table*}
\centering
\scalebox{0.73}{
\begin{tabular}{r | r | r | c c |c c|c c}
\hline
\multicolumn{3}{c}{} & \multicolumn{2}{c}{Predicate Classification} & \multicolumn{2}{c}{Scene Graph Classification} & \multicolumn{2}{c}{Scene Graph Detection} \\
\hline
Model & Fusion & Method & R@20~/~50~/~100 & mR@20~/~50~/~100 & R@20~/~50~/~100 & mR@20~/~50~/~100 & R@20~/~50~/~100 & mR@20~/~50~/~100  \\ 
\hline 
IMP+~\cite{xu2017scene, chen2019knowledge} & - & - & 52.7~/~59.3~/~61.3 & -~/~9.8~/~10.5 & 31.7~/~34.6~/~35.4 & -~/~5.8~/~6.0 & 14.6~/~20.7~/~24.5 & -~/~3.8~/~4.8  \\
FREQ~\cite{zellers2018neural, tang2019learning} & - & - & 53.6~/~60.6~/~62.2 & 8.3~/~13.0~/~16.0 & 29.3~/~32.3~/~32.9 & 5.1~/~7.2~/~8.5 & 20.1~/~26.2~/~30.1 & 4.5~/~6.1~/~7.1 \\
MOTIFS~\cite{zellers2018neural, tang2019learning} & - & - & 58.5~/~65.2~/~67.1 & 10.8~/~14.0~/~15.3 & 32.9~/~35.8~/~36.5 & 6.3~/~7.7~/~8.2 & 21.4~/~27.2~/~30.3 & 4.2~/~5.7~/~6.6  \\
KERN~\cite{chen2019knowledge} & - & - & -~/~65.8~/~67.6 & -~/~17.7~/~19.2 & -~/~36.7~/~37.4 & -~/~9.4~/~10.0 & -~/~27.1~/~29.8 & -~/~6.4~/~7.3  \\
VCTree~\cite{tang2019learning} & - & - & 60.1~/~66.4~/~68.1 & 14.0~/~17.9~/~19.4 & 35.2~/~38.1~/~38.8 & 8.2~/~10.1~/~10.8 & 22.0~/~27.9~/~31.3 & 5.2~/~6.9~/~8.0 \\
\hline 
\multirow{12}*{MOTIFS\textsuperscript{$\dagger$}} & \multirow{10}*{SUM} & Baseline & 59.5~/~66.0~/~67.9 & 11.5~/~14.6~/~15.8 & 35.8~/~39.1~/~39.9 & 6.5~/~8.0~/~8.5 & 25.1~/~32.1~/~36.9 & 4.1~/~5.5~/~6.8 \\ 
\cline{3-9}
& & Focal & 59.2~/~65.8~/~67.7 & 10.9~/~13.9~/~15.0 & 36.0~/~39.3~/~40.1 & 6.3~/~7.7~/~8.3 & 24.7~/~31.7~/~36.7 & 3.9~/~5.3~/~6.6 \\
& & Reweight & 45.4~/~57.0~/~61.7 & 16.0~/~20.0~/~21.9 & 24.2~/~29.5~/~31.5 & 8.4~/~10.1~/~10.9 & 18.3~/~24.4~/~29.3 & 6.5~/~8.4~/~9.8 \\
& & Resample & 57.6~/~64.6~/~66.7 & 14.7~/~18.5~/~20.0 & 34.5~/~37.9~/~38.8 & 9.1~/~11.0~/~11.8 & 23.2~/~30.5~/~35.4 & 5.9~/~8.2~/~9.7 \\
\cline{3-9}
& & X2Y & 58.3~/~65.0~/~66.9 & 13.0~/~16.4~/~17.6 & 35.2~/~38.6~/~39.5 & 6.9~/~8.6~/~9.2 & 24.8~/~32.1~/~36.7 & 5.1~/~6.9~/~8.1 \\
& & X2Y-Tr & 59.0~/~65.3~/~66.9 & 11.6~/~14.9~/~16.0 & 35.5~/~38.9~/~39.7 & 6.5~/~8.4~/~9.1 & 25.5~/~32.8~/~37.2 & 5.0~/~6.9~/~8.1 \\
\cline{3-9}
& & TE & 34.3~/~46.7~/~51.7 & 18.2~/~25.3~/~29.0 & 25.5~/~32.5~/~35.4 & 8.1~/~12.0~/~14.0 & 14.8~/~20.1~/~23.9 & 5.7~/~8.0~/~9.6 \\
& & NIE & 0.6~/~1.0~/~1.3 & 0.6~/~1.1~/~1.4 & 28.6~/~35.0~/~37.4 & 6.1~/~9.0~/~10.6 & 17.3~/~22.7~/~26.8 & 3.8~/~5.1~/~6.0\\
& & TDE & 33.6~/~46.2~/~51.4 & 18.5~/~25.5~/~29.1 & 21.7~/~27.7~/~29.9 & 9.8~/~13.1~/~14.9 & 12.4~/~16.9~/~20.3 & 5.8~/~8.2~/~9.8 \\
\cline{2-9}
& \multirow{2}*{GATE} & Baseline & 58.9~/~65.5~/~67.4 & 12.2~/~15.5~/~16.8 & 36.2~/~39.4~/~40.1 & 7.2~/~9.0~/~9.5 & 25.8~/~33.3~/~37.8 & 5.2~/~7.2~/~8.5 \\
& & TDE & 38.7~/~50.8~/~55.8 & 18.5~/~24.9~/~28.3 & 21.8~/~27.2~/~29.5 & 11.1~/~13.9~/~15.2 & 5.9~/~7.4~/~8.4 & 6.6~/~8.5~/~9.9 \\
\hline
\multirow{4}*{VTransE\textsuperscript{$\dagger$}} & \multirow{2}*{SUM} & Baseline & 59.0~/~65.7~/~67.6 & 11.6~/~14.7~/~15.8 & 35.4~/~38.6~/~39.4 & 6.7~/~8.2~/~8.7 & 23.0~/~29.7~/~34.3 & 3.7~/~5.0~/~6.0 \\
& & TDE & 36.9~/~48.5~/~53.1 & 17.3~/~24.6~/~28.0 & 19.7~/~25.7~/~28.5 & 9.3~/~12.9~/~14.8 & 13.5~/~18.7~/~22.6 & 6.3~/~8.6~/~10.5 \\
\cline{2-9}
& \multirow{2}*{GATE} & Baseline & 58.7~/~65.3~/~67.1 & 13.6~/~17.1~/~18.6 & 34.6~/~38.1~/~38.9 & 6.6~/~8.2~/~8.7 & 24.5~/~31.3~/~35.5 & 5.1~/~6.8~/~8.0 \\
& & TDE & 40.0~/~50.7~/~54.9 & 18.9~/~25.3~/~28.4 & 23.0~/~28.8~/~31.1 & 9.8~/~13.1~/~14.7 & 13.7~/~19.0~/~22.9 & 6.0~/~8.5~/~10.2 \\
\hline
\multirow{4}*{VCTree\textsuperscript{$\dagger$}} & \multirow{2}*{SUM} & Baseline & 59.8~/~66.2~/~68.1 & 11.7~/~14.9~/~16.1 & 37.0~/~40.5~/~41.4 & 6.2~/~7.5~/~7.9 & 24.7~/~31.5~/~36.2 & 4.2~/~5.7~/~6.9 \\
& & TDE & 36.2~/~47.2~/~51.6 & 18.4~/~25.4~/~28.7 & 19.9~/~25.4~/~27.9 & 8.9~/~12.2~/~14.0 & 14.0~/~19.4~/~23.2 & 6.9~/~9.3~/~11.1 \\
\cline{2-9}
& \multirow{2}*{GATE} & Baseline & 59.1~/~65.5~/~67.4 & 12.4~/~15.4~/~16.6 & 35.4~/~38.9~/~39.8 & 6.3~/~7.5~/~8.0 & 24.8~/~31.8~/~36.1 & 4.9~/~6.6~/~7.7\\
& & TDE & 39.1~/~49.9~/~54.5 & 17.2~/~23.3~/~26.6 & 22.8~/~28.8~/~31.2 & 8.9~/~11.8~/~13.4 & 14.3~/~19.6~/~23.3 & 6.3~/~8.6~/~10.3 \\
\hline
\hline
\end{tabular}
}
\caption{The SGG performances of Relationship Retrieval on both conventional \textbf{Recall@K} and \textbf{mean Recall@K}~\cite{tang2019learning, chen2019knowledge}. The SGG models reimplemented under our codebase are denoted by the superscript $\dagger$.}
\label{supp_tab:4}
\end{table*}

The full results of Relationship Retrieval, including both conventional Recall@K and the adopted mean Recall@K~\cite{tang2019learning, chen2019knowledge}, are given in Table~\ref{supp_tab:4}. Although a performance drop on conventional Recall@k is observed on TDE, the detailed analysis of the ``decreased'' predicates in Figure 6 of the original paper implies that it's caused by a more fine-grained predicate classification. 

\begin{figure}
   \begin{minipage}[b]{1\linewidth}
   \centerline{\includegraphics[width=85mm]{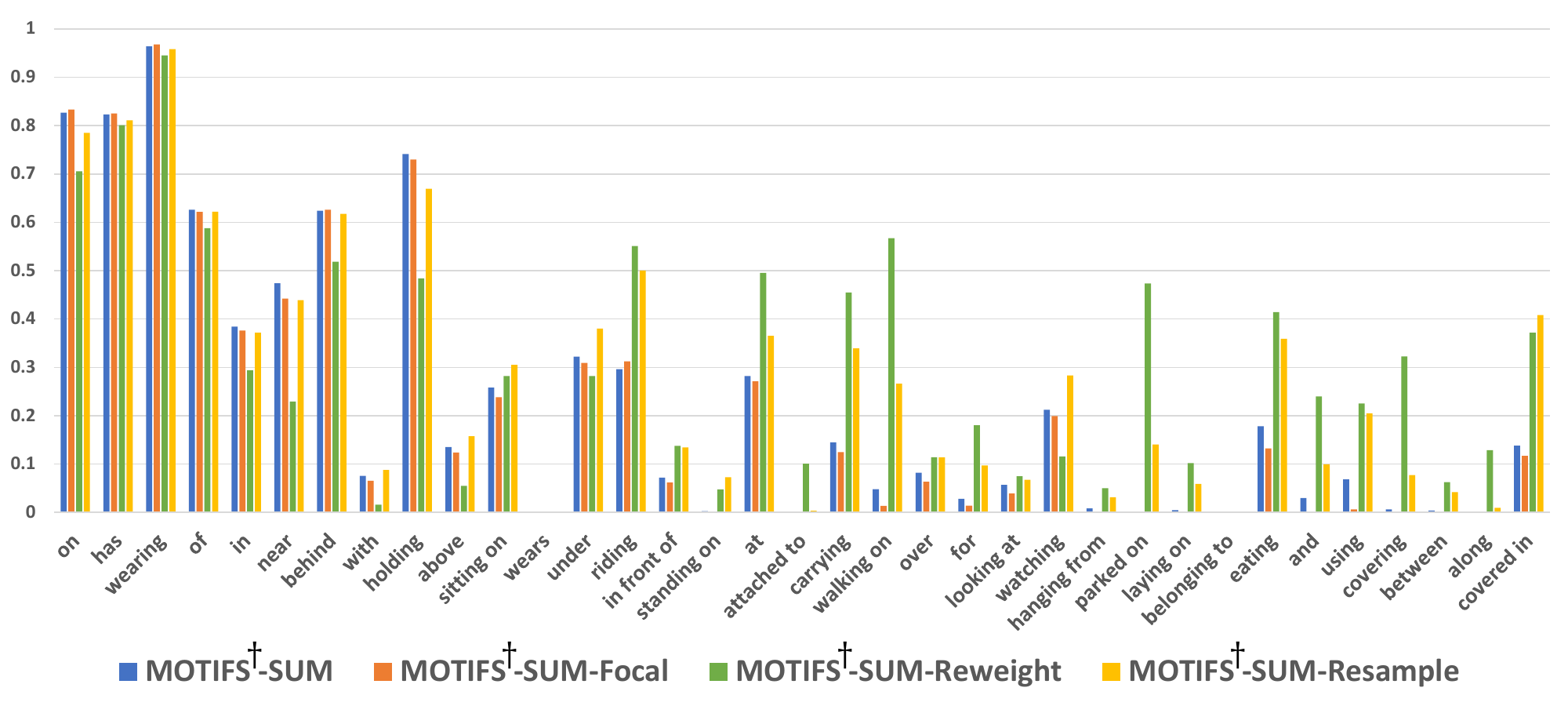}}
   \end{minipage}
   \caption{Conventional Debiasing Methods: Recall@100 on Predicate Classification for the most frequent 35 predicates.}
   \label{supp_fig:q4} 
\end{figure}

\begin{figure}
   \begin{minipage}[b]{1\linewidth}
   \centerline{\includegraphics[width=85mm]{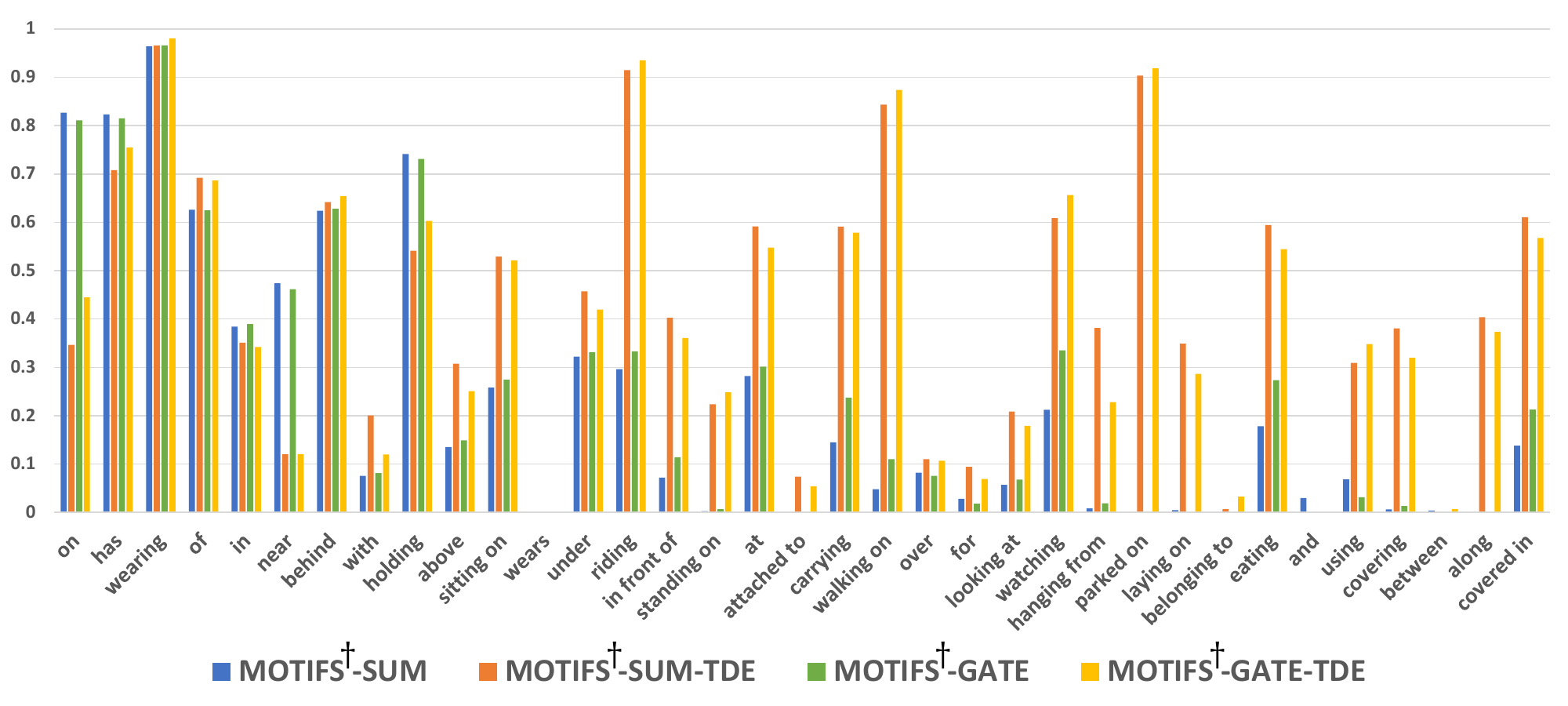}}
   \end{minipage}
   \caption{MOTIFS\textsuperscript{$\dagger$}~\cite{zellers2018neural}: Recall@100 on Predicate Classification for the most frequent 35 predicates.}
   \label{supp_fig:q1} 
\end{figure}

\begin{figure}
   \begin{minipage}[b]{1\linewidth}
   \centerline{\includegraphics[width=85mm]{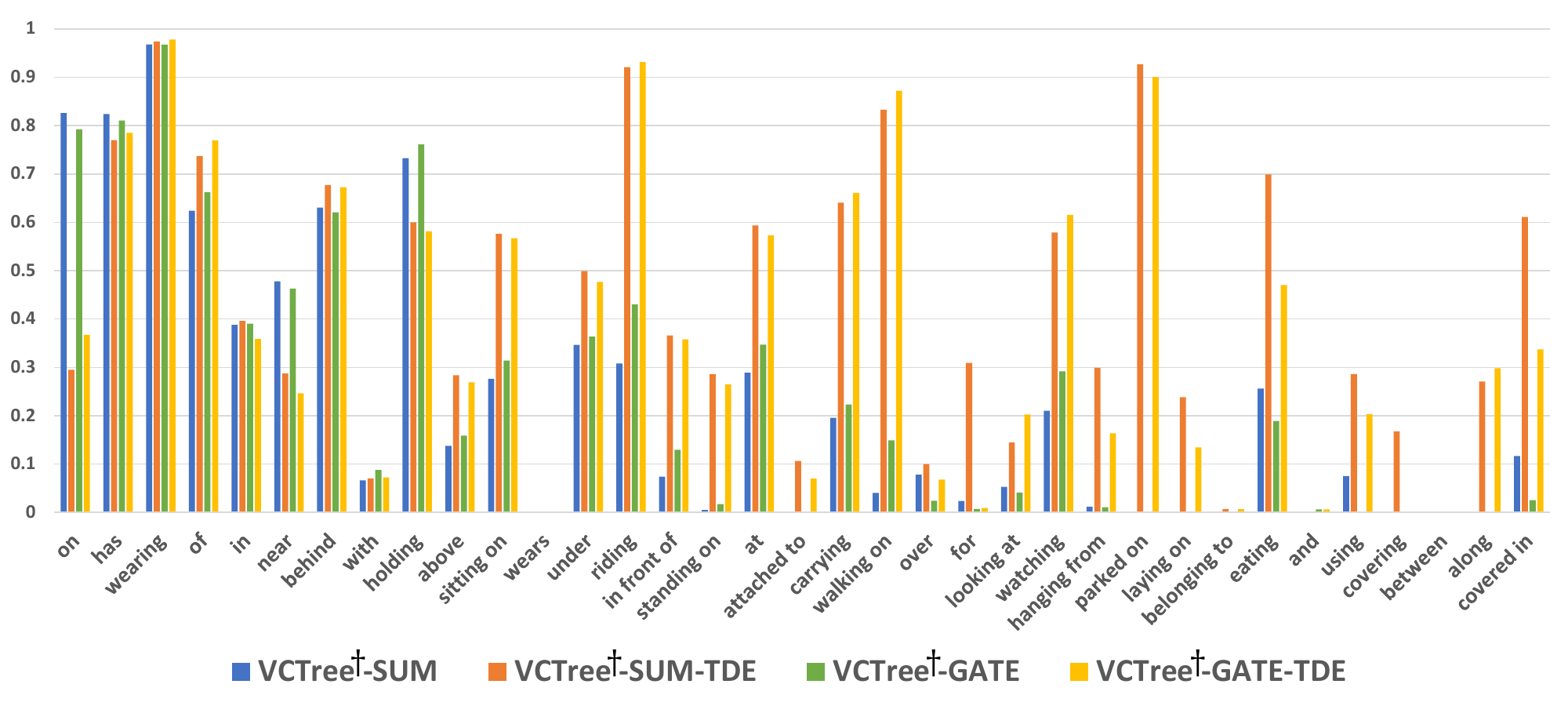}}
   \end{minipage}
   \caption{VCTree\textsuperscript{$\dagger$}~\cite{tang2019learning}: Recall@100 on Predicate Classification for the most frequent 35 predicates.}
   \label{supp_fig:q2} 
\end{figure}

\begin{figure}
   \begin{minipage}[b]{1\linewidth}
   \centerline{\includegraphics[width=85mm]{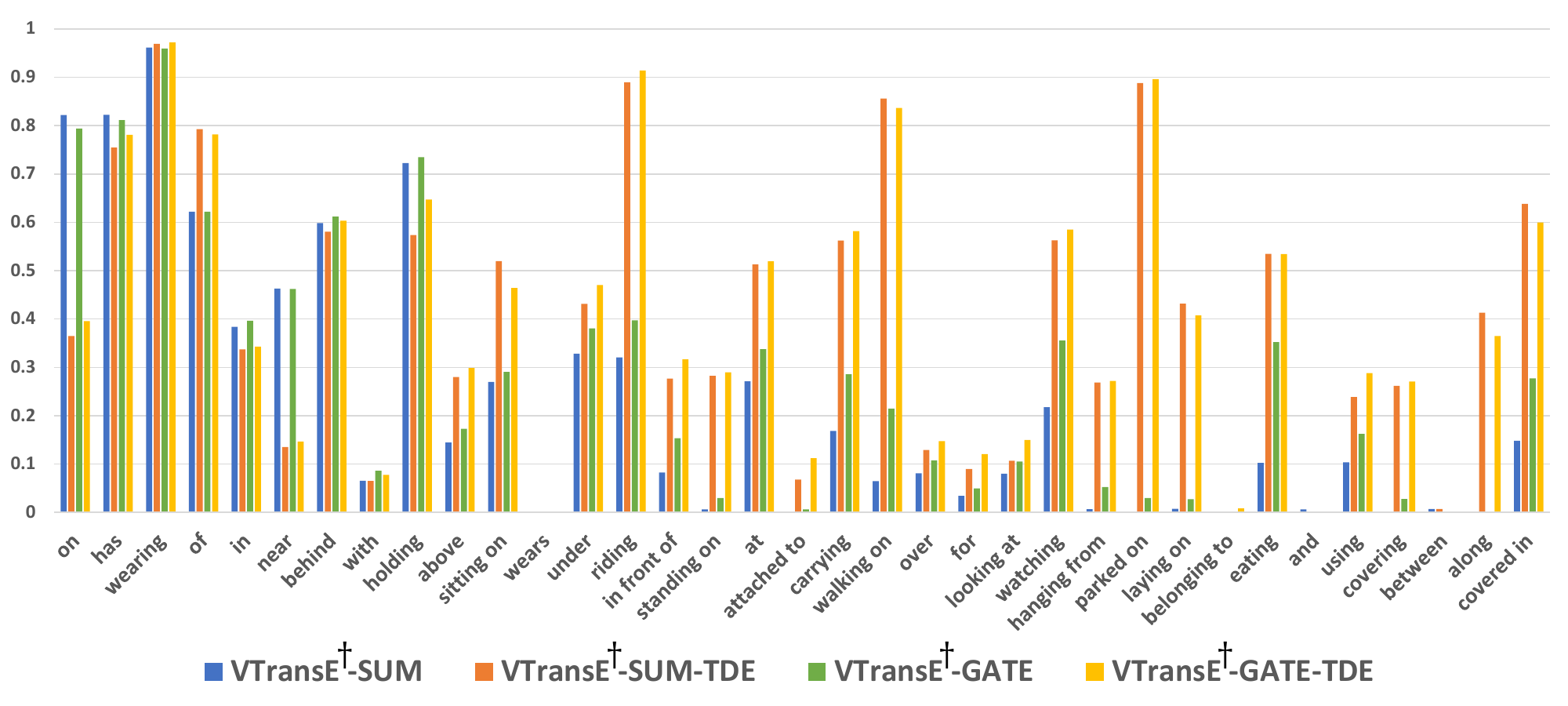}}
   \end{minipage}
   \caption{VTransE\textsuperscript{$\dagger$}~\cite{zhang2017visual}: Recall@100 on Predicate Classification for the most frequent 35 predicates.}
   \label{supp_fig:q3} 
\end{figure}

The detailed predicate-level Recall@100 on PredCls of all three models, two fusion functions and baseline \textit{vs.} TDE are given in Figure~\ref{supp_fig:q1}~\ref{supp_fig:q2}~\ref{supp_fig:q3}. Impressively, the distribution of the improved performances is no longer long-tailed while those conventional debiasing methods illustrated in Figure~\ref{supp_fig:q4} can't surpass the dataset distribution anyway. For TDE, very few decreased predicates are mainly due to the more fine-grained classification and we can observe significant improvements on their subclass predicates. Note that, unlike Reweight, which blindly hurt all frequent predicates, the proposed TDE will even improve some of the top-10 frequent predicates, like \texttt{behind} and \texttt{above}, which themselves are the subclasses of \texttt{near}. It further proves that the improvement of the proposed TDE doesn't come from hacking the distribution.

\section{Qualitative Studies}
\label{sec:qualitative}
\begin{figure*}
   \begin{minipage}[b]{1\linewidth}
   \centerline{\includegraphics[width=140mm]{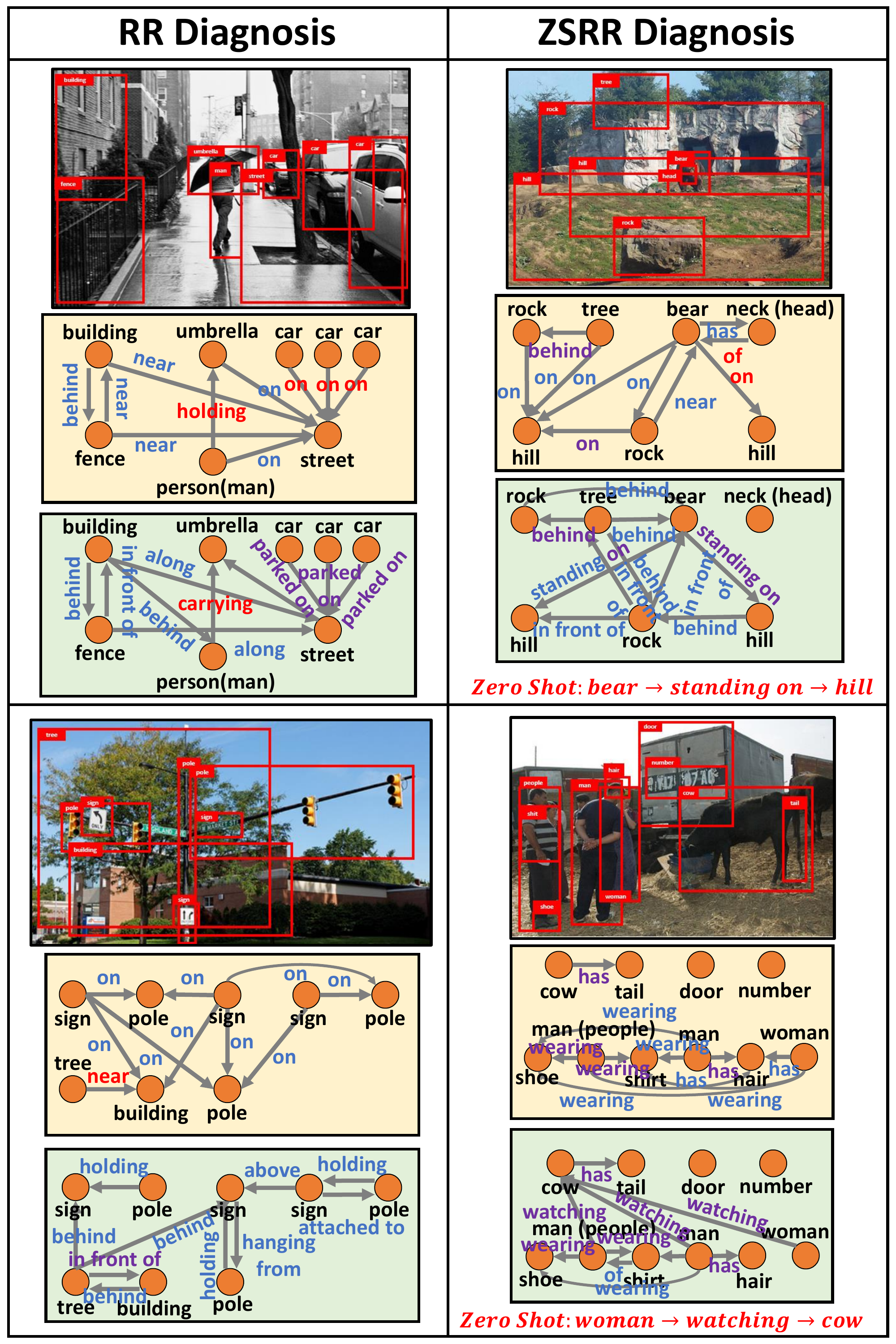}}
   \end{minipage}
   \caption{Top 10 Relationship Retrieval (RR) and Zero-Shot Relationship Retrieval (ZSRR) results of SGCls for MOTIFS\textsuperscript{$\dagger$}+SUM baseline (yellow box) and corresponding TDE (green box). The red predicates indicate misclassified relationships, the purple predicates are those correctly classified relationships (in ground truth), the blue predicates are those not labeled in ground truth.}
   \label{supp_fig:5} 
\end{figure*}

More Relationship Retrieval (RR) and Zero-Shot Relationship Retrieval (ZSRR) results are given in Figure~\ref{supp_fig:5}, where top 10 relationships under SGCls are selected for each image. As we can see, other than the trivial relationship problem, conventional baseline barely distinguishes different entities. For example, in the left bottom image, the same \texttt{sign} is almost \texttt{on} every \texttt{pole} in the baseline while the TDE results are more sensitive to different entities. However, one of the problem of TDE is that it over emphasizes the action predicates. It even uses \texttt{holding} for \texttt{pole} and \texttt{sign} while the predicate \texttt{on} used by the baseline is more natural in this case.

\begin{figure*}
   \begin{minipage}[b]{1\linewidth}
   \centerline{\includegraphics[width=140mm]{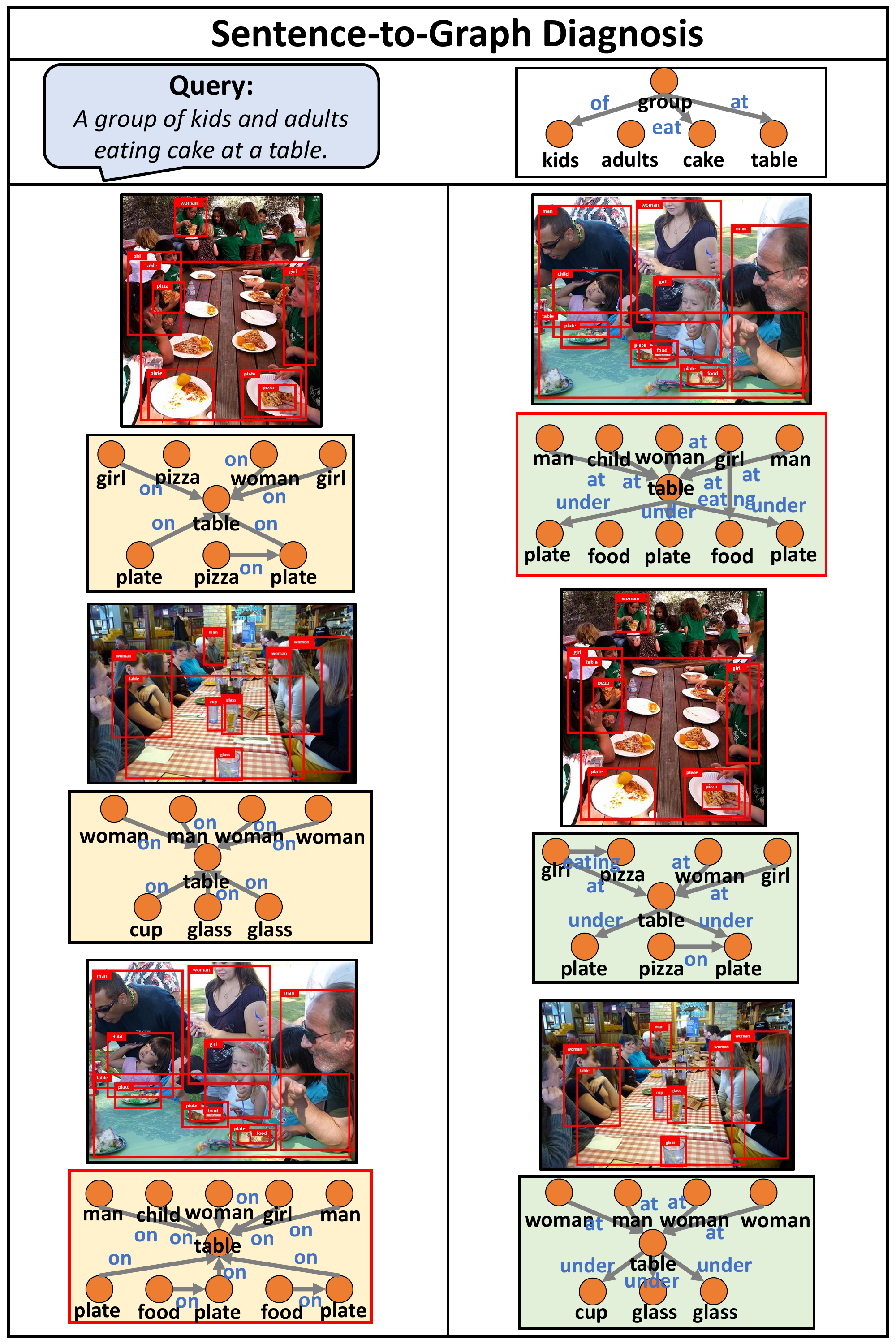}}
   \end{minipage}
   \caption{An example of Sentence-to-Graph Retrieval (S2GR) results for MOTIFS\textsuperscript{$\dagger$}+SUM baseline (yellow box) and corresponding TDE (green box). The red boxes indicate ground truth matching results. Note that we only draw sub-graphs containing important objects and predicates, because the original detected scene graphs from SGDet have too many trivial objects and predicates.}
   \label{supp_fig:6} 
\end{figure*}

Another example of Sentence-to-Graph Retrieval (S2GR) is illustrated in Figure~\ref{supp_fig:6}. Although we only reported sub-graphs of the original SGDet results, due to the limited space, we can still find that the conventional baseline model is not able to detect predicate like \texttt{eating}, which causes the detected SGs only provide the spatial relationships, missing the most discriminative word \texttt{eating} in the query caption.

\end{document}